\documentclass[11pt, logo, onecolumn, copyright, colorlinks=true, allcolors=blue]{nvidiatechreport}
\usepackage{graphicx} %
\usepackage{subcaption,ragged2e}
\usepackage[round]{natbib}
\usepackage[normalem]{ulem}
\usepackage[utf8]{inputenc} %
\usepackage[T1]{fontenc}    %
\usepackage{url}
\usepackage{tikz}
\usepackage{enumitem}
\usetikzlibrary{positioning}
\usetikzlibrary{
  arrows.meta,
  calc,
  positioning,
  fit,
  decorations.pathreplacing
}
\usepackage{array}
\usepackage{booktabs}
\usepackage{wrapfig}
\usepackage{caption} %
\usepackage{listings}
\usepackage{adjustbox}
\usepackage{footnote}
\usepackage{multirow}
\usepackage{amsmath}
\usepackage{amsfonts}
\usepackage{algorithm}
\usepackage{algpseudocode}
\usepackage{xspace}
\usepackage{breakcites}
\usepackage{blindtext}
\usepackage{svg}
\usepackage[most]{tcolorbox}
\usepackage{tabularx}
\usepackage{graphicx} 
\usepackage{xcolor}
\usepackage{float}
\usepackage{placeins}
\usepackage{needspace}
\usepackage{makecell}
\usepackage{colortbl}
\newcommand{\mtp}{MTP\xspace}
\newcommand{\hsm}{HSM\xspace}
\newcommand{\eagle}{EAGLE\xspace}
\newcommand{\eaglethree}{EAGLE-3\xspace}
\newcommand{\hsgd}{HSGD\xspace}
\newcommand{\ttt}{TTT\xspace}

\makeatletter
\DeclareRobustCommand{\super}{\@ifnextchar3{\nemotron@superthree}{Nemotron 3 Super\xspace}}
\def\nemotron@superthree3dot5{Nemotron 3.5 Super\xspace}
\makeatother
\newcommand{\lightning}{Nemotron 3.5 Lightning\xspace}
\newcommand{\ultra}{Nemotron 3 Ultra\xspace}

\title{Improving Tail-Augmented Speculative Drafting with Hidden State Mixing and Gradient Detach}

\author{
\textbf{Yaniv Galron\textsuperscript{*},
Rabeeh Karimi Mahabadi\textsuperscript{*},
Maor Ashkenazi\textsuperscript{*}}\authorcr
Roger Waleffe, Talor Abramovich, Tiyasa Mitra,
Yoav Miron, Dor Tsur,\authorcr

Sanjeev Satheesh, Jimmy Zhang, Jiaji Huang, Yonatan Geifman,\authorcr Mostofa Patwary, Mohammad Shoeybi, Bita Rouhani, Ran Zilberstein \\
{\large NVIDIA}\\
\textsuperscript{*}Equal contribution.
}
\date{}

\begin{document}

\begin{abstract}
\large \textbf{Abstract.}
\normalsize

Multi-token prediction (\mtp{}) modules and EAGLE-style drafters support efficient speculative decoding by reusing target-model hidden states. Their autoregressive use, however, introduces a training--inference mismatch: training supplies sequences of hidden states from the backbone or a single \mtp{} step, whereas inference attends to a target-model prefix followed by a growing tail of recurrently generated \mtp{} states. EAGLE-3's training-time test (\ttt{}) explicitly simulates this behavior during training, at the cost of additional memory and specialized attention masks.
We introduce two training-only methods for improving autoregressive drafting. Hidden State Mixing (\hsm{}) samples among backbone and \mtp{} hidden states aligned to the same prediction target, exposing the drafter to mixtures of states from different \mtp{} steps without explicitly simulating inference. Hidden State Gradient Detach (\hsgd{}) is a regularization method that stochastically blocks gradients through hidden state and embedding pathways while preserving the forward computation. Neither method adds inference cost.
In joint pretraining followed by supervised fine-tuning, \hsgd{} yields the highest acceptance rates at early draft positions, while \hsm{} performs best later in the draft. These acceptance rate trends favor \hsm{} for long-draft-length serving and \hsgd{} for short-draft-length serving. Furthermore, across four Nemotron models and DeepSeek-V4-Pro, \hsm{}-based \mtp{} boosting with a frozen backbone improves mean acceptance length by up to 52.2\% over the co-trained MTP. For EAGLE-3 drafters, \hsm{} offers competitive acceptance with \ttt{}, with 48\% lower peak reserved training memory in a four-step Qwen3-8B comparison. \hsm{} implementations are publicly available in Megatron-LM and NVIDIA Model Optimizer.

\end{abstract}

\maketitle

\begin{figure}[t]
  \centering
  \includegraphics[width=\linewidth]{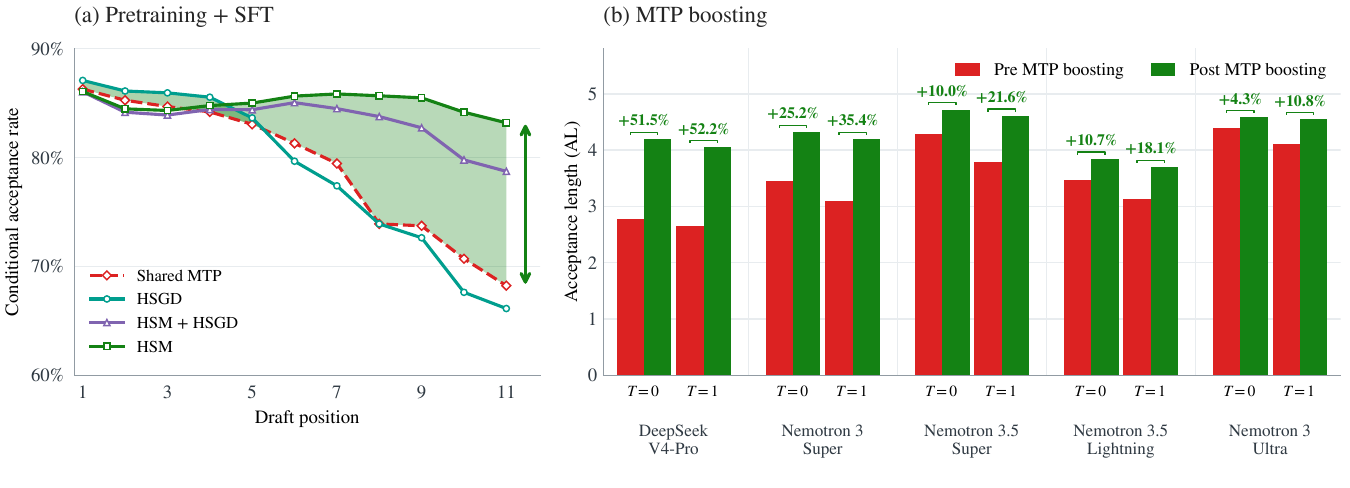}
  \caption{MTP training improvements.
  $T$ denotes the sampling temperature.
  (a) After joint pretraining and SFT, HSGD improves early-position
  acceptance rates, while HSM improves the draft tail. Mean acceptance length at draft
  length 11 increases from 5.19 with shared-parameter MTP to 5.57 with HSM
  ($T=0$). Results appear in
  Section~\ref{sec:pretraining-sft-experiments} and
  Appendix~\ref{app:pretraining-sft-results}.
  (b) HSM-based MTP boosting with a frozen backbone improves mean acceptance length across
  four Nemotron models and DeepSeek-V4-Pro.
  Bars show pre- and post-boosting AL at draft length seven;
  annotations report percentage gains relative to pre-boosting MTP. Results appear in Section~\ref{sec:hsm-boosting} and
  Appendix~\ref{app:boosting-al-by-draft-length}.
  Evaluation uses the SPEED-Bench qualitative split.}
  \label{fig:results-overview}
\end{figure}

\section{Introduction}
\label{sec:intro}
Large Language Model (LLM) inference is an inherently sequential process. During autoregressive decoding, the model generates one token per forward pass, with each new token becoming part of the input to the next pass. At low batch sizes, this process is often memory-bandwidth-bound: every decoding step reads the model weights and KV cache while performing relatively little computation. Speculative decoding shifts this balance by using a faster drafting mechanism to propose several future tokens and verifying them in a single target-model pass. Reusing the model weights and KV cache across multiple token positions increases arithmetic intensity, allowing more computation per byte transferred from memory. Rejection-sampling determines which tokens are accepted while preserving the target model's distribution, enabling lossless acceleration~\citep{leviathan2023fast}.

Drafting mechanisms range from simple approaches such as $n$-grams~\citep{saxena2023prompt} to separate, smaller LLMs, referred to as external draft models. Another family of methods, which we refer to as \emph{tail-augmented drafters}, attaches a lightweight drafting module to the target model and uses the target's hidden states as input. \eagle ~\citep{li2024eagle} and \mtp (Multi-Token Prediction) ~\citep{gloeckle2024better} generate draft tokens autoregressively using a shallow decoder module, whereas parallel drafters such as DFlash~\citep{chen2026dflash} and LiLiCorr~\citep{rusanovsky2026lilicorr} predict a block of tokens in a single pass. Generating multiple tokens in one pass reduces drafting overhead,
making parallel drafting particularly attractive for smaller models
and low-concurrency serving, where longer drafts can be verified
with modest additional latency. At higher concurrency, shorter drafts
are often preferable to limit verification cost, reducing the advantage
of parallel block generation and making lightweight autoregressive
\mtp more attractive. Autoregressive \mtp is also widely shipped with
modern models as a built-in drafting mechanism co-trained with the
backbone. We study how to train these modules for effective speculative
decoding.

MTP was originally proposed as an auxiliary training objective for improving the target model accuracy. By predicting several future tokens from each position, the additional supervision can improve representation learning and downstream model quality. The resulting \mtp modules have also been used as speculative drafters~\citep{gloeckle2024better, deepseekai2025deepseekv3technicalreport}. Because these modules can be trained jointly with the target throughout pretraining and post-training, they have the potential to become stronger drafters than tail-augmented drafters trained only after the target model has been frozen. Nevertheless, relatively little work has studied how \mtp should be trained specifically for high acceptance rates. 
We examine this limitation in Section~\ref{sec:frozen-backbone-mtp-training} through using the released DeepSeek-V4-Pro~\citep{deepseekai2026deepseekv4} \mtp module, which exhibits a substantial drop in acceptance rates after the first drafted token.
This drop is consistent with its pretraining setup, which used only one MTP step. DeepSeek has since
released DSpark~\citep{cheng2026dspark}, a semi-autoregressive drafter designed for speculative decoding.

A central difficulty in drafting with autoregressive MTPs is that they encounter different inputs during training and inference. We identify two forms of this training--inference mismatch. The first arises when separate \mtp modules are trained for different prediction steps but a single module is later reused autoregressively. Sharing parameters across \mtp steps addresses this mismatch. A second, deeper mismatch remains even with parameter sharing. During training, the drafter is conditioned on either hidden states produced from the target, or hidden states produced from a previous MTP step. During inference, it is conditioned on its own generated hidden states, while its KV cache contains mostly target-generated hidden states, and an increasing number of recursively generated \mtp states. The resulting distribution shift grows with draft depth and particularly harms acceptance rates at later positions. \eaglethree~\citep{li2026eagle} addresses this problem through training-time test (TTT), which explicitly simulates autoregressive draft generation during training. Although effective, TTT requires specialized attention masks and additional unrolled drafting steps. Its memory costs therefore grow with the number of simulated steps (Section~\ref{sec:eagle3-alternative}), making it harder to scale and inconvenient to integrate into large-scale pretraining.

We introduce \emph{Hidden State Mixing} (\hsm), a lightweight alternative
to TTT that mixes aligned hidden states from different \mtp steps,
exposing the drafter to varied contexts without explicitly simulating
autoregressive inference during training.
We also introduce \emph{Hidden State Gradient Detach} (\hsgd), a
regularization approach motivated by the coupling between successive \mtp steps:
later prediction losses also update earlier representations.
\hsgd randomly blocks gradients through selected hidden state and
embedding pathways while preserving the forward computation.
Both methods operate only during training and add no inference cost.

Our experiments show that the benefits depend on both the training regime
and draft position. During joint pretraining followed by supervised
fine-tuning, \hsgd improves acceptance rates at early draft positions,
while \hsm improves the draft tail
(Figure~\ref{fig:results-overview}(a)). These profiles suggest different
serving targets: shorter drafts are often preferable at high concurrency,
where early-position acceptance rates matter most, while longer drafts
can be more useful at low concurrency, where higher tail acceptance rates
become valuable.
\hsm also benefits frozen-backbone \emph{MTP boosting}: continued training
of the MTP module improves mean acceptance length across four Nemotron
models and DeepSeek-V4-Pro (Figure~\ref{fig:results-overview}(b)). We further
evaluate \hsm as a lightweight alternative to TTT for EAGLE-3 drafters
(Section~\ref{sec:eagle3-alternative}).

\textbf{Contributions.} This paper makes the following contributions:
\begin{itemize}[leftmargin=*]
    \item We formalize the training--inference mismatch in autoregressive \mtp drafting and distinguish the parameter-sharing mismatch from the deeper hidden state exposure mismatch.
    \item We introduce \hsm, a simple training-only method that injects structured hidden state variation by mixing aligned states across \mtp steps.
    \item We introduce \hsgd, a method that regularizes gradient flow through the recurrent \mtp steps.
    \item We evaluate both methods in frozen-backbone training and co-training (pretraining followed by supervised fine-tuning), and compare \hsm with TTT in EAGLE-style training.
    \item We show that \hsm and \hsgd improve different draft positions, motivating future exploration of serving-aware choices of the \mtp training strategy.
    \item We provide practical guidance for co-training MTP from backbone pretraining through supervised fine-tuning.
\end{itemize}

\newpage
\section{Background: Tail-Augmented Speculative Drafting}
\label{sec:background}

Tail-augmented speculative decoding uses a lightweight drafter conditioned on hidden states from the target model. Reusing these representations simplifies the prediction task, allowing a small module, often comprising one or a few decoder blocks, to propose future tokens with substantially less computation than a separate draft model.

Medusa~\citep{medusa} introduced a simple approach to tail-augmented drafting by attaching a separate prediction head for each future-token position. Its prediction horizon was therefore tied to the number of trained heads. Moreover, all heads predicted from the same target-model hidden state before the next token had been sampled. Different sampling outcomes lead to different future trajectories, leaving the heads uncertain about which trajectory to predict.

EAGLE~\citep{li2024eagle} addressed this ambiguity by conditioning feature prediction on both the preceding hidden states and the embeddings of the sampled tokens. The sampled token grounds the next hidden state prediction in a specific continuation. Unlike Medusa, EAGLE uses a full Transformer decoder block, applied autoregressively to generate successive hidden states and tokens. This recurrent computation allows generation to extend beyond a fixed set of position-specific prediction heads.

A similar design is used in co-trained multi-token prediction (MTP), where the auxiliary MTP loss jointly trains draft modules and the target backbone~\citep{gloeckle2024better,deepseekai2025deepseekv3technicalreport}. This loss was initially motivated by the richer training signal and resulting improvements in backbone quality; however, the draft modules can also serve as built-in speculative drafters. In~\cite{gloeckle2024better}, separate draft modules independently consume the same backbone representations. DeepSeek-V3's MTP formulation, which we adopt in this work, also uses position-specific modules, but connects them sequentially, combining the preceding hidden states with shifted token embeddings at each stage. This token-conditioned computation resembles EAGLE, although EAGLE reuses a single decoder block autoregressively. As opposed to a post-trained EAGLE, in co-trained MTP, the auxiliary MTP loss additionally shapes the backbone representations.

Candidate continuations can also be organized into token trees and verified in parallel, as demonstrated by SpecInfer~\citep{miao2023specinfer}. EAGLE adopts tree-structured generation, and EAGLE-2~\citep{li2025eagle} extends it with dynamic trees whose expansion depends on drafter confidence. This strategy can also be applied to MTP modules and other autoregressive drafters.
Complementary approaches, including DFlash~\citep{chen2026dflash}, DSpark~\citep{cheng2026dspark}, and LiLiCorr~\citep{rusanovsky2026lilicorr}, support generating multiple candidate tokens in parallel. Here, we focus on \emph{autoregressive} tail-augmented drafters.

Autoregressive draft token generation also introduces a training--inference challenge: the drafter must consume its own predicted hidden states during generation. EAGLE tackled this with a feature-matching loss, alongside its token-prediction objective, encouraging predicted states to remain close to the corresponding target-model states. EAGLE-3~\citep{li2026eagle} removes this feature-matching constraint and introduces training-time test (TTT), which simulates successive drafting steps during training. Later predictions are consequently exposed to contexts combining target-derived representations with recursively generated drafter states. We examine this training--inference mismatch in Section~\ref{sec:mismatch} and subsequently propose an alternative training approach.

\newpage
\section{The \mtp Training-Inference Mismatch}
\label{sec:mismatch}

We focus on two training--inference mismatches that arise when using a single MTP module for autoregressive drafting (Figure~\ref{fig:mtp-mismatch}). The first concerns the recurrent reuse of modules trained with position-specific modules (i.e., each MTP step uses its own set of parameters). Prior work~\citep{cai2025fastmtp} addresses this mismatch through parameter sharing across draft steps. The second persists even with shared parameters: the hidden state context encountered during drafting differs from that used during training. We distinguish these two effects below, motivating HSM introduced in Section~\ref{sec:hsm}.
A further mismatch arises from conditioning on reference tokens during training and generated tokens at inference, motivating HSGD introducted in section ~\ref{sec:hsgd}. This is common to autoregressive models trained with teacher forcing and is not specific to tail-augmented drafters.

\subsection{Position-Specific Training and Recurrent Drafting}
\label{sec:basic-mismatch}

With position-specific parameters, each prediction step has its own MTP module. The first consumes target-model hidden states, while subsequent modules consume representations from the preceding module.
At inference, the trained modules can be applied once in sequence, or one module can be reused recurrently (Figure~\ref{fig:mtp-mismatch}(a)). Applying each module once limits the draft length to the number of trained modules. Recurrent reuse removes this limit, but changes the module's input distribution: the first MTP module trained on target-model hidden states will later consume its own generated hidden states during inference, although this behavior was absent from training.

Sharing parameters across prediction steps during training addresses this architectural mismatch, as demonstrated by FastMTP~\citep{cai2025fastmtp}. The same module is trained on representations produced by its earlier applications (Figure~\ref{fig:mtp-mismatch}(b)). Recent models using this approach include \super{}~\citep{nvidia2026nemotron3superopen}, \ultra{}~\citep{nvidia2026nemotron3ultra}, and \lightning{}~\citep{nvidia2026nemotron35}. Qwen3-Next~\citep{qwen2025qwen3next}, GLM-5.2~\citep{glm5team2026glm5}, and Gemma-4~\citep{gemmateam2026gemma4} also perform well when their MTP module is reused recurrently at inference.

\begin{figure}[t]
  \centering
  \definecolor{mtpTarget}{HTML}{72B7B2}
\definecolor{mtpStepOne}{HTML}{F2A65A}
\definecolor{mtpStepTwo}{HTML}{A985B4}
\definecolor{mtpStepThree}{HTML}{7087C6}
\definecolor{mtpInk}{HTML}{263442}
\definecolor{mtpBlue}{HTML}{2F6B9A}
\definecolor{mtpRed}{HTML}{C6534C}
\definecolor{mtpGreen}{HTML}{4B8F6A}
\definecolor{mtpGray}{HTML}{72808E}
\definecolor{mtpPanel}{HTML}{F5F7F9}
\definecolor{mtpWarmPanel}{HTML}{FCF7F5}
\definecolor{mtpCoolPanel}{HTML}{F3F8FA}

\begin{tikzpicture}[
  x=1cm,y=1cm,
  font=\fontfamily{ptm}\selectfont,
  text=mtpInk,
  >={Latex[length=2.2mm,width=1.5mm]},
  paneltitle/.style={font=\bfseries\small, anchor=west},
  subtitle/.style={font=\bfseries\scriptsize, anchor=west},
  tinytext/.style={font=\fontsize{6.0}{6.8}\selectfont, align=center},
  module/.style={draw=mtpBlue!80!black, fill=mtpBlue, text=white,
    rounded corners=4pt, minimum width=0.76cm, minimum height=0.56cm,
    inner sep=1.5pt, line width=0.8pt, font=\fontsize{5.7}{6.3}\selectfont\bfseries},
  modulegray/.style={module, draw=mtpGray!85!black, fill=mtpGray},
  moduleorange/.style={module, draw=mtpStepOne!85!black, fill=mtpStepOne!88!black},
  modulepurple/.style={module, draw=mtpStepTwo!85!black, fill=mtpStepTwo!90!black},
  modulegreen/.style={module, draw=mtpGreen!85!black, fill=mtpGreen},
  seq/.style={draw=mtpInk!48, fill=white, rounded corners=1.5pt,
    minimum width=0.76cm, minimum height=0.42cm, inner sep=0.6pt,
    line width=0.55pt, font=\fontsize{5.5}{6.2}\selectfont},
  smallstate/.style={draw=mtpInk!48, fill=white, circle, minimum size=0.38cm,
    inner sep=0.2pt, line width=0.55pt, font=\fontsize{5.4}{6.0}\selectfont},
  option/.style={draw=mtpGray!60, fill=mtpGray!7, rounded corners=2pt,
    minimum width=1.00cm, minimum height=0.34cm, inner sep=1pt,
    line width=0.6pt, font=\fontsize{5.8}{6.4}\selectfont\bfseries,
    text=mtpGray},
  state/.style={draw=mtpInk!45, rounded corners=1.5pt,
    minimum width=0.82cm, minimum height=0.48cm, inner sep=1.2pt,
    line width=0.55pt, font=\tiny},
  targetstate/.style={state, fill=mtpTarget!34, draw=mtpTarget!80!black},
  stepone/.style={state, fill=mtpStepOne!38, draw=mtpStepOne!85!black},
  steptwo/.style={state, fill=mtpStepTwo!35, draw=mtpStepTwo!85!black},
  flow/.style={-{Latex[length=1.25mm,width=0.85mm]}, draw=mtpInk!68,
    line width=0.72pt, shorten <=0.5pt, shorten >=1.2pt},
]

\fill[mtpPanel, rounded corners=5pt] (0,6.95) rectangle (16.4,10.10);
\node[paneltitle] at (0.28,9.80) {(a) Position-specific MTP modules};
\draw[mtpInk!13, line width=0.65pt] (6.82,7.22) -- (6.82,9.42);

\node[subtitle, text=mtpBlue] at (0.45,9.25) {Training};
\node[seq, minimum width=0.70cm] (tr0) at (0.48,8.35) {$h_{0:t}^{0}$};
\node[modulegray, minimum width=0.70cm] (tm1) at (1.45,8.35) {MTP$_1$};
\node[seq, minimum width=0.70cm] (tr1) at (2.42,8.35) {$h_{0:t}^{1}$};
\node[moduleorange, minimum width=0.70cm] (tm2) at (3.39,8.35) {MTP$_2$};
\node[seq, minimum width=0.70cm] (tr2) at (4.36,8.35) {$h_{0:t}^{2}$};
\node[modulepurple, minimum width=0.70cm] (tm3) at (5.33,8.35) {MTP$_3$};
\node[seq, minimum width=0.70cm] (tr3) at (6.30,8.35) {$h_{0:t}^{3}$};
\foreach \a/\b in {tr0/tm1,tm1/tr1,tr1/tm2,tm2/tr2,tr2/tm3,tm3/tr3}
  \draw[flow] (\a) -- (\b);

\node[subtitle, text=mtpRed] at (7.12,9.25) {Inference};
\node[option] at (7.55,8.68) {Option A};
\node[smallstate] (fa0) at (8.45,8.68) {$h$};
\node[modulegray] (fm1) at (9.28,8.68) {MTP$_1$};
\node[moduleorange] (fm2) at (10.42,8.68) {MTP$_2$};
\node[modulepurple] (fm3) at (11.56,8.68) {MTP$_3$};
\foreach \a/\b in {fa0/fm1,fm1/fm2,fm2/fm3}
  \draw[flow] (\a) -- (\b);
\node[draw=mtpRed, fill=white, circle, minimum size=0.36cm, inner sep=0pt,
  font=\scriptsize\bfseries, text=mtpRed] (finiteStop) at (12.48,8.68) {$\times$};
\draw[flow, draw=mtpRed] (fm3) -- (finiteStop);
\node[anchor=west, font=\tiny\bfseries, text=mtpRed, align=left]
  at (12.82,8.68) {draft length $\leq$\\number of MTP modules};

\node[option] at (7.55,7.58) {Option B};
\node[smallstate] (ra0) at (8.45,7.58) {$h$};
\node[modulegray] (rm1a) at (9.28,7.58) {MTP$_1$};
\node[smallstate, fill=mtpRed!5, draw=mtpRed!65] (ra1) at (10.11,7.58) {$\hat h^1$};
\node[modulegray] (rm1b) at (10.94,7.58) {MTP$_1$};
\node[smallstate, fill=mtpRed!5, draw=mtpRed!65] (ra2) at (11.77,7.58) {$\hat h^2$};
\node[modulegray] (rm1c) at (12.60,7.58) {MTP$_1$};
\node[smallstate, fill=mtpRed!5, draw=mtpRed!65] (ra3) at (13.43,7.58) {$\hat h^3$};
\draw[flow] (ra0) -- (rm1a);
\draw[flow] (rm1a) -- (ra1);
\draw[-{Latex[length=1.25mm,width=0.85mm]}, draw=mtpRed, dashed,
  line width=0.75pt, shorten <=0.5pt, shorten >=1.2pt] (ra1) -- (rm1b);
\draw[flow] (rm1b) -- (ra2);
\draw[-{Latex[length=1.25mm,width=0.85mm]}, draw=mtpRed, dashed,
  line width=0.75pt, shorten <=0.5pt, shorten >=1.2pt] (ra2) -- (rm1c);
\draw[flow] (rm1c) -- (ra3);
\node[anchor=west, font=\tiny\bfseries, text=mtpRed] at (13.82,7.58)
  {unseen self-inputs};

\fill[mtpCoolPanel, rounded corners=5pt] (0,4.64) rectangle (16.4,6.72);
\node[paneltitle] at (0.28,6.44) {(b) Parameter sharing};
\node[seq] (sh0) at (4.30,5.55) {$h_{0:t}^{0}$};
\node[modulegreen] (sm1) at (5.48,5.55) {MTP};
\node[seq] (sh1) at (6.66,5.55) {$h_{0:t}^{1}$};
\node[modulegreen] (sm2) at (7.84,5.55) {MTP};
\node[seq] (sh2) at (9.02,5.55) {$h_{0:t}^{2}$};
\node[modulegreen] (sm3) at (10.20,5.55) {MTP};
\node[seq] (sh3) at (11.38,5.55) {$h_{0:t}^{3}$};
\node[font=\small, text=mtpGray] (sdots) at (12.38,5.55) {$\cdots$};
\foreach \a/\b in {sh0/sm1,sm1/sh1,sh1/sm2,sm2/sh2,sh2/sm3,sm3/sh3,sh3/sdots}
  \draw[flow] (\a) -- (\b);
\node[draw=mtpGreen, rounded corners=2pt, fill=mtpGreen!8, text=mtpGreen,
  font=\tiny\bfseries, minimum width=1.15cm, minimum height=0.30cm,
  inner xsep=4pt, inner ysep=1pt] (sharedtheta) at (7.84,4.88)
  {$\theta_{\mathrm{MTP}}$};
\draw[draw=mtpGreen!88, dashed, line width=0.78pt]
  (sm1.south) -- (5.48,5.12) -- (10.20,5.12) -- (sm3.south);
\draw[draw=mtpGreen!88, dashed, line width=0.78pt]
  (sm2.south) -- (7.84,5.12) -- (sharedtheta.north);

\fill[mtpWarmPanel, rounded corners=5pt] (0,0) rectangle (16.4,4.42);
\node[paneltitle] at (0.28,4.14) {(c) Hidden state exposure mismatch};
\draw[mtpInk!13, line width=0.65pt] (7.65,0.46) -- (7.65,3.76);

\node[subtitle, text=mtpBlue] at (0.52,3.60) {Training};
\node[anchor=east, font=\tiny\bfseries, text=mtpGray] at (1.72,3.08) {MTP 1};
\node[anchor=east, font=\tiny\bfseries, text=mtpGray] at (1.72,2.28) {MTP 2};
\node[anchor=east, font=\tiny\bfseries, text=mtpGray] at (1.72,1.48) {MTP 3};
\foreach \x/\lab in {2.18/$h_0^0$,3.18/$h_1^0$,4.18/$h_2^0$,5.18/$\cdots$,6.18/$h_t^0$}
  \node[targetstate] at (\x,3.08) {\lab};
\foreach \x/\lab in {2.18/$h_0^1$,3.18/$h_1^1$,4.18/$h_2^1$,5.18/$\cdots$,6.18/$h_t^1$}
  \node[stepone] at (\x,2.28) {\lab};
\foreach \x/\lab in {2.18/$h_0^2$,3.18/$h_1^2$,4.18/$h_2^2$,5.18/$\cdots$,6.18/$h_t^2$}
  \node[steptwo] at (\x,1.48) {\lab};

\node[subtitle, text=mtpRed] at (8.50,3.60) {Inference};
\node[anchor=east, font=\tiny\bfseries, text=mtpGray] at (9.78,3.08) {MTP prefill};
\node[anchor=east, font=\tiny\bfseries, text=mtpGray] at (9.78,2.28) {MTP draft 1};
\node[anchor=east, font=\tiny\bfseries, text=mtpGray] at (9.78,1.48) {MTP draft 2};
\foreach \x/\lab in {10.25/$h_0^0$,11.15/$h_1^0$,12.05/$\cdots$,12.95/$h_t^0$}
  \node[targetstate] at (\x,3.08) {\lab};
\foreach \x/\lab in {10.25/$h_0^0$,11.15/$h_1^0$,12.05/$\cdots$,12.95/$h_t^0$}
  \node[targetstate] at (\x,2.28) {\lab};
\node[stepone] at (13.85,2.28) {$h_t^1$};
\foreach \x/\lab in {10.25/$h_0^0$,11.15/$h_1^0$,12.05/$\cdots$,12.95/$h_t^0$}
  \node[targetstate] at (\x,1.48) {\lab};
\node[stepone] at (13.85,1.48) {$h_t^1$};
\node[steptwo] at (14.75,1.48) {$h_t^2$};
\draw[decorate, decoration={brace, amplitude=4pt, mirror},
  draw=mtpRed, line width=0.8pt]
  (13.40,1.06) -- (15.20,1.06)
  node[midway, below=4pt, font=\scriptsize\bfseries, text=mtpRed]
  {mixed-depth tail};
\draw[->, draw=mtpRed, line width=1.0pt] (13.28,3.34) -- (15.62,3.34)
  node[midway, above, font=\tiny\bfseries, text=mtpRed] {growing mismatch};

\pgfresetboundingbox
\path[use as bounding box] (-0.05,-0.05) rectangle (16.45,10.15);
\end{tikzpicture}
  
  \input{figures/main_figure/mtp_mismatch_figure_caption.tex}
  \label{fig:mtp-mismatch}
\end{figure}

\subsection{Hidden State Exposure Mismatch} 
\label{sec:exposure-mismatch}

Even with parameter sharing, a hidden state exposure mismatch remains
(Figure~\ref{fig:mtp-mismatch}(c)). During training, each MTP step
processes a sequence of hidden states from a single depth: either
target-model hidden states or outputs of a preceding MTP step. During inference,
the MTP KV cache initially represents a target-model prefix, and successive
drafting steps append entries computed from recursively generated MTP
steps. From the second draft position onward, the module therefore
attends to a context combining target-model hidden states with hidden states from
different MTP steps. This composition is absent from standard training
and can compound hidden state prediction errors as drafting proceeds.

The released DeepSeek-V4-Pro MTP module provides a concrete example of
later-position degradation. On the SPEED-Bench~\citep{speedbenchnvidia} qualitative split at
$T=0$ and draft length seven, the acceptance rate falls from $84\%$
at the first position to $63\%$ at the second and $45\%$ at the third
(Figure~\ref{fig:mtp-boosting-ar}(a);
Table~\ref{tab:mtp-boosting-ar}). This illustrates the challenge
of maintaining acceptance rate during recurrent drafting, although it does
not isolate the contribution of hidden state exposure mismatch.

EAGLE~\citep{li2024eagle} sought to limit feature drift by combining
its token-prediction objective with a feature-matching loss. Its one-step prediction is supervised against the corresponding
target-model hidden state:
\[
\mathcal{L}_{\mathrm{feat}}
=
\operatorname{SmoothL1}
\left(\mathbf{h}_{t-1}^1,\mathbf{h}_{t}^0\right),
\]
where $\mathbf{h}_{t-1}^1$ is the MTP output and
$\mathbf{h}_{t}^0$ is the target-model hidden state aligned with the same
prediction target, $x_{t+1}$. This objective encourages generated states
to remain close to target-model representations, but does not explicitly
reproduce the mixed-depth contexts encountered during drafting.
EAGLE-3 instead exposes the drafter to these contexts through
training-time test (TTT), discussed next.

\subsection{Why \ttt Helps and Why It Is Expensive}
\label{sec:ttt-cost}
EAGLE-3's training-time test (\ttt)~\citep{li2026eagle}
trains the drafter on the hidden state contexts it encounters during
inference. Starting from each target-model prefix, it feeds the
drafter's outputs back into the same module for several steps. At
step $d$, the context combines target-model states $\mathbf{h}_i^0$
with preceding drafter outputs $\mathbf{h}_t^r$:
\[
\mathcal{C}_t^d =
\bigl[\mathbf{h}_0^0,\ldots,\mathbf{h}_t^0,\;
      \mathbf{h}_t^1,\ldots,\mathbf{h}_t^{d-1}\bigr].
\]
Shifted token embeddings are supplied at each step.
TTT runs these continuations from all $N$ training positions in
parallel. Each round adds $N$ states, so at round $d$ the $N$ queries
have $dN$ candidate keys. The attention mask lets each continuation
use its target prefix and its own preceding states. A dense
implementation therefore uses a mask of size $N\times dN$; sparse
implementations can exploit its structure. Further rounds also add
activations and KV states, so the overall cost depends on the
implementation and attention backend.

In our Llama-3.1-8B-Instruct comparison, TTT trained with four steps shows
declining acceptance rates beyond position four
(Figure~\ref{fig:eagle-main-ar-t0}(a)). However, increasing the number of training steps is costly: in the ModelOpt Qwen3-8B setup, increasing from four to eight steps raises peak reserved memory per GPU from 70.3 to 197.2\,GiB
(Figure~\ref{fig:ttt-memory}(f)). This motivates our proposed Hidden State Mixing
(\hsm), introduced next. %

\section{Method 1: Hidden State Mixing}
\label{sec:hsm}
This section introduces \hsm, a tail-augmented drafter training algorithm that aims to reduce the training--inference mismatch in hidden state context (Section~\ref{sec:exposure-mismatch}) by injecting variation from mixing aligned hidden states at different MTP steps into the drafter's hidden state inputs during training. We first describe standard MTP training and its inference-time KV cache evolution, highlighting this mismatch. We then introduce aligned hidden states and show how \hsm changes the hidden state source used by this computation. Finally, we interpret \hsm as structured noise injection, where the structure comes from mixing states that predict the same target token.

\textbf{Definitions.}
Let \(\mathbf{e}_i=E(x_i)\) denote the embedding of the token \(x_i\).
For two vectors \(\mathbf{h},\mathbf{e}\in\mathbb{R}^{D}\),
let \(\mathbf{h}\oplus\mathbf{e}\in\mathbb{R}^{2D}\) denote their concatenation
along the hidden dimension.
Additionally let \(\mathbf{h}_t^0\) denote the target-model's hidden state at position \(t\),
and let \(\mathbf{h}_t^d\), for \(d\geq1\), denote the hidden state produced by the \(d\)-th
MTP step. For a token sequence \((x_1,\ldots,x_T)\) during training,
we compute \(\mathbf{h}_t^d\) only at positions satisfying
\(t+d+1\leq T\), where the ground-truth token \(x_{t+d+1}\)
exists.

\subsection{Standard MTP Training}
\label{sec:hsm-baseline}
During standard teacher-forced MTP training,
the first MTP step combines the target-model hidden state sequence with the one-token-shifted ground-truth token embeddings.
The first step therefore computes
\[
\bigl[
  \mathbf{h}_0^0 \oplus \mathbf{e}_1,\; \ldots,\;
  \mathbf{h}_t^0 \oplus \mathbf{e}_{t+1}
\bigr]
\;\xrightarrow{\text{MTP module}}\;
\bigl[\mathbf{h}_0^1,\; \ldots,\; \mathbf{h}_t^1\bigr].
\]
The next MTP step directly consumes this output sequence and pairs it with
embeddings shifted by one additional position:
\[
\bigl[
  \mathbf{h}_0^1 \oplus \mathbf{e}_2,\; \ldots,\;
  \mathbf{h}_t^1 \oplus \mathbf{e}_{t+2}
\bigr]
\;\xrightarrow{\text{MTP module}}\;
\bigl[\mathbf{h}_0^2,\; \ldots,\; \mathbf{h}_t^2\bigr].
\]
More generally, the input at position \(t\) to MTP step
\(d+1\) is \(\mathbf{h}_t^d \oplus \mathbf{e}_{t+d+1}\), and the
corresponding output is \(\mathbf{h}_t^{d+1}\).
Figure~\ref{fig:mtp-mismatch}(c) illustrates this training-time evolution,
in which each MTP step consumes a sequence from a single hidden state depth.
Consequently, \(\mathbf{h}_t^d\) is trained to predict \(x_{t+d+1}\).
HSM retains the same MTP steps and prediction targets but changes the hidden state source supplied to each step during training.

\subsection{Inference-Time MTP KV Cache}
\label{sec:hsm-cache}

During inference, the \mtp module maintains its own KV cache,
separate from the target model's cache. At the start of a drafting
phase, this cache is populated from target-model hidden states.
The first draft step uses this target-derived context. From the
second step onward, each new cache entry is computed from the
preceding MTP step's hidden state and the embedding of its selected
token. The cache therefore contains a target-derived prefix followed
by a growing tail of MTP-generated states. For a drafting phase starting from
$\bigl[\mathbf{h}_0^0,\ldots,\mathbf{h}_t^0\bigr]$,
the effective input sequence at step $d$ is
\[
\bigl[
  \mathbf{h}_0^0 \oplus \mathbf{e}_1,\; \ldots,\;
  \mathbf{h}_t^0 \oplus \mathbf{e}_{t+1},\;
  \bigl(\mathbf{h}_t^r \oplus \tilde{\mathbf{e}}_{t+r+1}\bigr)_{r=1}^{d-1}
\bigr]
\;\xrightarrow{\text{MTP module}}\;
\bigl[ \mathbf{h}_t^d \bigr].
\]

Previously processed inputs are represented by cached keys and values.
Here, $\mathbf{e}_{t+1}$ embeds the token selected by the target model,
while $\tilde{\mathbf{e}}_{t+r+1}=E(\tilde{x}_{t+r+1})$ embeds the token
selected at MTP step $r\geq1$, using the same embedding function $E$
as in training.
As illustrated in Figure~\ref{fig:mtp-mismatch}(c), standard training
supplies each MTP step with a hidden state sequence from a single
depth, whereas inference combines a target-model prefix with states
from successive MTP steps. This mixed-depth context motivates \hsm.

\subsection{Aligned Hidden States}
\label{sec:hsm-alignment}
\begin{figure}[t]
  \centering
  \resizebox{\linewidth}{!}{\definecolor{hsmtarget}{HTML}{76B7B2}
\definecolor{hsmstepone}{HTML}{F28E2B}
\definecolor{hsmsteptwo}{HTML}{B07AA1}
\definecolor{hsmblue}{HTML}{2F6B9A}
\definecolor{hsmlight}{HTML}{F4F7FA}
\definecolor{hsmtext}{HTML}{263442}

\begin{tikzpicture}[
  font=\sffamily,
  text=hsmtext,
  state/.style={draw, rounded corners=2pt, minimum width=1.22cm,
    minimum height=0.62cm, inner sep=2pt, line width=0.7pt, font=\small},
  source/.style={state, minimum width=1.42cm},
  flow/.style={->, >=stealth, line width=0.9pt, color=hsmblue},
  candidate/.style={->, >=stealth, line width=0.75pt, dashed, color=hsmblue!75},
  paneltitle/.style={font=\bfseries\small, anchor=west},
  note/.style={font=\scriptsize, align=center}
]

\fill[hsmlight, rounded corners=5pt] (-1.55,-0.08) rectangle (7.65,5.30);
\fill[hsmlight, rounded corners=5pt] (7.95,-0.08) rectangle (16.55,5.30);
\draw[hsmblue!22, line width=0.8pt] (7.80,0.18) -- (7.80,5.00);

\node[paneltitle] at (-0.25,4.92) {(a) Aligned hidden states};
\node[paneltitle] at (8.20,4.92) {(b) Hidden State Mixing during training};

\node[anchor=east, font=\scriptsize\bfseries] at (1.15,3.50) {Target, $d=0$};
\node[anchor=east, font=\scriptsize\bfseries] at (1.15,2.25) {MTP step, $d=1$};
\node[anchor=east, font=\scriptsize\bfseries] at (1.15,1.00) {MTP step, $d=2$};

\node[font=\tiny, color=hsmtext!85] at (2.00,4.08) {position $t-3$};
\node[font=\tiny, color=hsmtext!65] at (3.50,4.08) {$t-2$};
\node[font=\tiny, color=hsmtext!65] at (5.00,4.08) {$t-1$};
\node[font=\tiny, color=hsmtext!65] at (6.50,4.08) {$t$};

\node[state, fill=hsmtarget!24] (h0t)  at (2.00,3.50) {$h_{t-3}^0$};
\node[state, fill=hsmtarget!24] (h0t1) at (3.50,3.50) {$h_{t-2}^0$};
\node[state, fill=hsmtarget!45, draw=hsmblue, line width=1.2pt] (h0t2) at (5.00,3.50) {$h_{t-1}^0$};
\node[state, fill=hsmtarget!24] (h0t3) at (6.50,3.50) {$h_{t}^0$};

\node[state, fill=hsmstepone!22] (h1t)  at (2.00,2.25) {$h_{t-3}^1$};
\node[state, fill=hsmstepone!42, draw=hsmblue, line width=1.2pt] (h1t1) at (3.50,2.25) {$h_{t-2}^1$};
\node[state, fill=hsmstepone!22] (h1t2) at (5.00,2.25) {$h_{t-1}^1$};

\node[state, fill=hsmsteptwo!42, draw=hsmblue, line width=1.2pt] (h2t) at (2.00,1.00) {$h_{t-3}^2$};
\node[state, fill=hsmsteptwo!22] (h2t1) at (3.50,1.00) {$h_{t-2}^2$};

\node[source, fill=hsmtarget!45, draw=hsmblue] (s0) at (8.90,3.45) {$h_{t-1}^{0}$};
\node[source, fill=hsmstepone!42, draw=hsmblue] (s1) at (8.90,2.25) {$h_{t-2}^{1}$};
\node[source, fill=hsmsteptwo!42, draw=hsmblue] (s2) at (8.90,1.05) {$h_{t-3}^{2}$};
\node[note, anchor=south] at (8.90,3.88) {Aligned sources\\$\mathcal{A}_{t}^{\le 2}$};

\node[draw=hsmblue, rounded corners=5pt, fill=white, minimum width=1.55cm,
  minimum height=1.05cm, align=center, font=\scriptsize\bfseries] (selector)
  at (11.15,2.25) {Sampler};
\draw[candidate] (s0.east) -- (selector.west);
\draw[candidate] (s1.east) -- (selector.west);
\draw[candidate] (s2.east) -- (selector.west);

\node[state, fill=hsmstepone!42, draw=hsmblue, line width=1.2pt,
  minimum width=1.45cm] (mixed) at (13.15,2.25) {$\widetilde h_{t-1}$};
\draw[flow] (selector.east) -- node[above, font=\scriptsize] {} (mixed.west);

\node[draw=hsmblue, rounded corners=4pt, fill=hsmblue!10, minimum width=1.30cm,
  minimum height=1.10cm, align=center, font=\scriptsize\bfseries] (block)
  at (15.20,2.25) {ordinary\\MTP module};
\draw[flow] (mixed.east) -- (block.west);

\node[state, fill=hsmtarget!24, minimum width=0.65cm, minimum height=0.36cm]
  at (9.00,0.18) {};
\node[anchor=west, font=\tiny] at (9.42,0.18) {target state};
\node[state, fill=hsmstepone!22, minimum width=0.65cm, minimum height=0.36cm]
  at (11.50,0.18) {};
\node[anchor=west, font=\tiny] at (11.92,0.18) {MTP step 1};
\node[state, fill=hsmsteptwo!22, minimum width=0.65cm, minimum height=0.36cm]
  at (14.05,0.18) {};
\node[anchor=west, font=\tiny] at (14.47,0.18) {MTP step 2};

\end{tikzpicture}}
  \caption{Hidden State Mixing (HSM).
  (a) The blue outlines identify the aligned hidden state set
  $\{h^0_{t-1},h^1_{t-2},h^2_{t-3}\}$: all three states predict $x_t$,
  but originate from different MTP steps and therefore carry different accumulated
  noise.
  (b) During training, HSM samples one available aligned source
  $\widetilde h_{t-1}\in\mathcal{A}_t^{\leq d}$. The selected hidden state supplies the query, key, and value projections of the ordinary MTP
  block. HSM changes only the training-time hidden state. Inference is unchanged.}
  \label{fig:hsm-overview}
\end{figure}
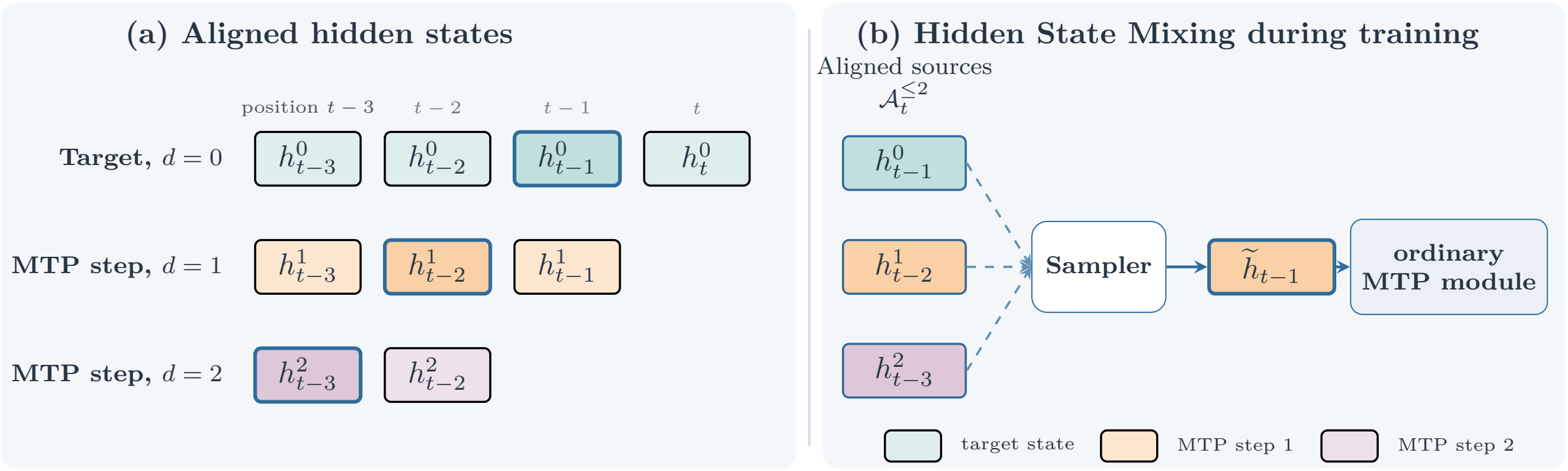

We call hidden states from different MTP steps
\emph{aligned} if they are trained to predict the same token. For example,
\[
\mathbf{h}^0_{t-1}, \qquad \mathbf{h}^1_{t-2}, \qquad \mathbf{h}^2_{t-3}
\]
are aligned because all three predict \(x_t\). Figure~\ref{fig:hsm-overview} illustrates
this alignment. More generally, the states aligned with
\(\mathbf{h}_{t-1}^0\) are
\[
\mathcal{A}_t^{\leq d}
=
\left\{
\mathbf{h}_{t-r-1}^{r}
\;\middle|\;
r \in \{0,\ldots,d\}
\right\}.
\]
Every state in \(\mathcal{A}_t^{\leq d}\) predicts
\(x_{t}\).

\subsection{Algorithm}
\label{sec:hsm-implementation}

\label{sec:hsm-mixing-strategies}

HSM constructs each MTP step's input sequence by sampling aligned hidden
states from the backbone and preceding MTP steps. We describe two distinct approaches to this sampling procedure: uniform layer sampling, our default strategy, and exponential sparse replacement (HSM-ESR).

\paragraph{Uniform layer sampling.}
We maintain a list of hidden state sequences from the backbone and all
previously computed MTP steps. Before each MTP step, these sequences are
aligned so that the states at a given position correspond to the same
prediction target (Section~\ref{sec:hsm-alignment}). For each token position,
we independently sample one of the available aligned states uniformly to
construct the input hidden state sequence. The hidden states produced by
the current \mtp step are appended to the list for subsequent steps.
We use this strategy by default and denote it by \hsm in the experiments.

After computing MTP step $d$, HSM constructs the hidden state
input to step $d+1$. For each token position $t$, it samples
from the available aligned states that predict $x_t$: $\widetilde{\mathbf{h}}_{t-1}\in \mathcal{A}_t^{\leq d}$.

The states in this set have already been computed; consequently, they may
themselves reflect \hsm choices made at earlier \mtp steps. HSM leaves the ground-truth embedding $\mathbf{e}_t$ unchanged and replaces only the hidden state source. Since those are aligned hidden states, the ground-truth embeddings match and no change is needed. The selected source is used for all three
attention projections (omitting embeddings for simplicity):
\[
\mathbf{q}_t = W_Q \widetilde{\mathbf{h}}_{t-1},\quad
\mathbf{k}_t = W_K \widetilde{\mathbf{h}}_{t-1},\quad
\mathbf{v}_t = W_V \widetilde{\mathbf{h}}_{t-1}.
\]
Thus HSM changes only the hidden state seen by the attention block. The sampler is used only
during training, while inference retains the ordinary sequential MTP computation.
Let \(J_{d,t}\) index the hidden state sampled from \(\mathcal{A}_t^{\leq d}\). The token-wise selection under uniform layer sampling is:
\[
J_{d,t} \sim \operatorname{Uniform}\{0,\ldots,d\},\qquad
\widetilde{\mathbf{h}}_{t-1}
= \mathbf{h}^{J_{d,t}}_{t-1-J_{d,t}}.
\]

Algorithm~\ref{alg:frozen-mtp-hsm} provides pseudocode for \hsm with uniform layer sampling and a frozen backbone.

\begin{algorithm}[t]
\caption{Frozen-Backbone MTP Forward during Training with Hidden State Mixing}
\label{alg:frozen-mtp-hsm}
\begin{algorithmic}[1]

\Require Token sequence \(x_{1:T}\)
\Require Frozen Backbone model \(B\)
\Require MTP drafter \(M\)
\Require \(D\) MTP steps
\Require Frozen token embedding \(E\)
\Require Depth-aligned training labels \(\mathcal{Y}^{(1:D)}\)

\Statex
\State \textbf{// Frozen-backbone forward pass}
\State \(H^{(0)} \gets \Call{StopGradient}{B(x_{1:T})}\)
    \Comment{Target-model hidden state sequence}
\State \(\mathcal{H} \gets [H^{(0)}]\)
\State \(u \gets x_{1:T}\)
    \Comment{Ground-truth token sequence, shifted once per MTP step}
\State \(\mathcal{L}_{\mathrm{MTP}} \gets 0\)

\Statex
\State \textbf{// Feed input to the MTP D times}
\For{\(d \gets 1\) \textbf{to} \(D\)}
    \State \(\mathcal{H}  \gets
        \Call{Align}{\mathcal{H}}\)
        \Comment{Shift states to a common prediction target}
    \State \(\widetilde{H}^{(d)} \gets
        \Call{Sample}{\mathcal{H}}\)
        \Comment{Independently sample one source per token}
    \State \(u \gets \Call{ShiftLeft}{u}\) \Comment{Align ground-truth token inputs with the sampled states}
    \State \(H^{(d)} \gets M\!\left(\widetilde{H}^{(d)}, E(u)\right)\)
        \Comment{MTP hidden state sequence at depth \(d\)}
    \State \(\mathcal{H}.\Call{Append}{H^{(d)}\!}\)
    \State \(\mathcal{L}_{\mathrm{MTP}} \gets
        \Call{UpdateLoss}{\mathcal{L}_{\mathrm{MTP}},\; H^{(d)},\; \mathcal{Y}^{(d)}}\)
    
\EndFor

\end{algorithmic}
\end{algorithm}

\paragraph{Exponential sparse replacement (HSM-ESR).}
ESR starts from the target-model hidden state sequence. After each
MTP step, it replaces a random subset of positions with aligned
outputs from that step, while the remaining positions retain their
current states. The updated sequence then serves as input to the
next MTP step. The number of replacements decreases exponentially
with depth, so later steps preserve more of the existing mixture.
This produces inputs containing states from several MTP steps
without resampling the source at every position.

\paragraph{Implementation.}
Uniform layer sampling is implemented in Megatron-LM,\footnote{\href{https://github.com/NVIDIA/Megatron-LM/blob/af8eef5e3e3fc3698b062162ca7356f825c99c8f/megatron/core/transformer/multi_token_prediction.py\#L107}{Megatron-LM HSM implementation.}}
and both variants are implemented in NVIDIA Model Optimizer
(ModelOpt) for EAGLE-style draft models.\footnote{\href{https://github.com/NVIDIA/Model-Optimizer/blob/27a504479bc0a615fadade6e8a9759041d778f0b/modelopt/torch/speculative/plugins/hf_eagle.py\#L819}{ModelOpt HSM implementation.}}

\subsection{Interpretation as Structured Noise Injection}
\label{sec:hsm-noise}

An MTP step can be viewed as advancing a hidden state by one token
while introducing approximation noise:
\[
\mathbf{h}_{t-2}^1
=
\mathbf{h}_{t-1}^0+\boldsymbol{\eta}_t,
\]
where both states predict $x_t$. 
For inference robustness, one training-time option would be to perturb MTP-generated states with noise sampled from a normal or uniform distribution. HSM instead uses the difference between aligned states as a structured perturbation. For the two states above, applying this correction at randomly selected positions gives 
\[
\begin{aligned}
\widetilde{\mathbf{h}}_{t-1}
&=
\mathbf{h}_{t-1}^0
+m_t\bigl(\mathbf{h}_{t-2}^1 - \mathbf{h}_{t-1}^0\bigr)
=
\mathbf{h}_{t-1}^0
+m_t\boldsymbol{\eta}_t,
\end{aligned}
\]
where $m_t\in\{0,1\}$ indicates whether the perturbation is applied.
Selected positions use aligned MTP-generated states, while unselected
positions retain backbone states. Thus, adding this structured
perturbation at random positions is equivalent to hidden state mixing. Sampling from the larger set
$\mathcal{A}_t^{\leq d}$ extends this interpretation to differences
between states from additional MTP steps.

\newpage
\section{Method 2: Hidden State Gradient Detach}
\label{sec:hsgd}
  Hidden State Mixing modifies the forward-pass inputs of the MTP modules during training.
  We introduce \emph{Hidden State Gradient Detach} (HSGD), a complementary
  training-only regularizer that instead modifies the backward flow through
  the recurrent MTP computation. HSGD leaves all forward activations
  unchanged while stochastically blocking selected upstream gradient paths.
  Every MTP step retains its prediction loss and continues to train the
  parameters used by that step.
  We apply gradient detachment to two paths: the hidden state connection
  between consecutive MTP steps and the token-embedding input to each stage.
  Figure~\ref{fig:hsgd-overview} illustrates these two detachment paths.

 \subsection{Motivation}
  \label{sec:hsgd-motivation}

  Sequential MTP training couples prediction objectives across draft
  positions. To make this dependency explicit, first consider standard
  sequential MTP without detachment and with position-specific parameters
  $\theta_j$. Let $\mathcal{L}_{d}$ denote the prediction loss at MTP step
  $d$. Because the hidden state produced at one stage becomes an input to
  the next, $\mathcal{L}_{d}$ backpropagates through every preceding MTP
  stage:
  \begin{equation}
      \frac{\mathrm{d}\mathcal{L}_{d}}
           {\mathrm{d}\theta_j}
      =
      \frac{\partial \mathcal{L}_{d}}
           {\partial \mathbf{h}^{d}}
      \left(
          \frac{\partial \mathbf{h}^{d}}
               {\partial \mathbf{h}^{d-1}}
          \cdots
          \frac{\partial \mathbf{h}^{j+1}}
               {\partial \mathbf{h}^{j}}
      \right)
      \frac{\partial \mathbf{h}^{j}}
           {\partial \theta_j},
      \qquad j \leq d.
      \label{eq:mtp-cross-stage-gradient}
  \end{equation}
  Here $\mathbf{h}^{k}$ denotes the full hidden state sequence, with its
  entries flattened for Jacobian notation. The Jacobians are multiplied
  in descending stage order, and their product is the identity when
  $j=d$. The local derivative
  $\partial\mathbf{h}^{j}/\partial\theta_j$ holds the stage input fixed.
  Consequently, an early MTP step is optimized not only for its own
  next-draft-token objective, but also to produce representations that
  facilitate every subsequent MTP loss. When a single MTP module is
  recurrently reused across draft positions, its shared parameters receive
  the sum of the gradient contributions associated with all unrolled
  applications. These include both direct contributions from each
  application's prediction loss and indirect contributions through its
  influence on subsequent applications. For a loss $\mathcal{L}_d$,
  the shared-parameter gradient is obtained by evaluating
  Equation~\eqref{eq:mtp-cross-stage-gradient} for each application
  $j\leq d$ at the tied parameter values and summing the contributions.

  During teacher-forced training, the first MTP step consumes target-model
  hidden states, while subsequent steps consume states produced by
  preceding MTP applications. Each step also receives ground-truth
  future-token embeddings. An early representation is therefore optimized
  both for its own prediction objective and for its usefulness to later
  steps. These objectives may encourage co-adaptation across steps under
  teacher-forced conditioning, motivating regularization of the paths
  through which later losses modify earlier representations.

  As briefly mentioned in Section~\ref{sec:mismatch}, there is a related mismatch in the embedding pathway that is true for autoregressive models in general. During standard
  training, MTP step $d$ receives the ground-truth embedding
    \[
      \mathbf{e}_{t+d}
      =
      E\left(x_{t+d}\right),
   \]
  whereas autoregressive drafting may instead provide
  \[
      \widehat{\mathbf{e}}_{t+d}
      =
      E\left(\widehat{x}_{t+d}\right),
      \qquad
      \widehat{x}_{t+d} \neq x_{t+d}.
  \]
  When the embedding table is jointly trained with the MTP objectives,
  gradients from the teacher-forced MTP pathway can encourage the
  embedding representation to co-adapt with the drafter under clean
  future-token inputs. We hypothesize that regularizing this gradient
  pathway can reduce embedding--drafter co-adaptation under teacher-forced
  conditioning.
  
  These observations motivate regularizing how credit is assigned across
  the MTP computation graph. Rather than always allowing every loss to
  update the complete sequence of upstream states and embeddings, we
  stochastically truncate selected gradient paths. Each stage must still
  solve the same prediction problem from exactly the same forward
  activations, but downstream losses cannot consistently coordinate all
  earlier components of the chain. We hypothesize that this reduces
  cross-step co-adaptation. The resulting changes in acceptance rates can differ
  across draft positions; the co-training experiments with HSGD alone show
  gains concentrated at the first few positions.

  HSGD does not by itself correct the
  forward hidden state distribution shift. Instead, it acts as a
  backward-only regularizer for cross-stage and embedding--drafter
  gradient coupling under teacher-forced conditioning. It can be used
  with HSM, which modifies the forward
  hidden state inputs, without introducing training-time inference
  simulation.

\begin{figure}[t]
  \centering
  \definecolor{hsgdInk}{HTML}{263442}
\definecolor{hsgdBlue}{HTML}{2F6B9A}
\definecolor{hsgdRed}{HTML}{C6534C}
\definecolor{hsgdGreen}{HTML}{4B8F6A}
\definecolor{hsgdPurple}{HTML}{8F6A9D}
\definecolor{hsgdOrange}{HTML}{E59A4A}
\definecolor{hsgdTeal}{HTML}{72B7B2}
\definecolor{hsgdGray}{HTML}{72808E}
\definecolor{hsgdCool}{HTML}{F3F8FA}
\definecolor{hsgdWarm}{HTML}{FCF7F5}

\begin{tikzpicture}[
  x=1cm,y=1cm,
  font=\fontfamily{ptm}\selectfont,
  text=hsgdInk,
  >={Latex[length=2.2mm,width=1.5mm]},
  paneltitle/.style={font=\bfseries\small, anchor=west},
  module/.style={draw=hsgdPurple!85!black, fill=hsgdPurple, text=white,
    rounded corners=4pt, minimum width=1.02cm, minimum height=0.62cm,
    inner sep=2pt, line width=0.85pt, font=\tiny\bfseries},
  state/.style={draw=hsgdTeal!75!black, fill=hsgdTeal!24,
    rounded corners=2pt, minimum width=0.82cm, minimum height=0.52cm,
    inner sep=1pt, line width=0.65pt, font=\tiny},
  embed/.style={draw=hsgdOrange!85!black, fill=hsgdOrange!24,
    rounded corners=2pt, minimum width=0.90cm, minimum height=0.52cm,
    inner sep=1pt, line width=0.65pt, font=\tiny},
  loss/.style={draw=hsgdInk!55, fill=white, circle, minimum size=0.62cm,
    inner sep=0pt, line width=0.65pt, font=\tiny\bfseries},
  flow/.style={->, draw=hsgdBlue, line width=0.82pt,
    shorten <=1.0pt, shorten >=2.0pt},
  back/.style={->, draw=hsgdRed, dashed, line width=0.92pt},
  faintback/.style={draw=hsgdRed!22, dashed, line width=0.75pt},
  tinylabel/.style={font=\fontsize{6.1}{6.8}\selectfont, align=center},
]

\fill[hsgdCool, rounded corners=5pt] (0,2.55) rectangle (16.4,6.00);
\node[paneltitle] at (0.28,5.70)
  {(a) Hidden state detachment};

\node[state] (h0) at (0.85,3.92) {$\mathbf{h}_t^0$};
\node[module] (m1) at (2.15,3.92) {MTP};
\node[state] (h1) at (3.45,3.92) {$\mathbf{h}_t^1$};
\node[module] (m2) at (5.40,3.92) {MTP};
\node[state] (h2) at (6.70,3.92) {$\mathbf{h}_t^2$};
\node[module] (m3) at (8.65,3.92) {MTP};
\node[state] (h3) at (9.95,3.92) {$\mathbf{h}_t^3$};
\node[loss] (l3) at (11.28,3.92) {$\mathcal L_3$};
\foreach \a/\b in {h0/m1,m1/h1,h1/m2,m2/h2,h2/m3,m3/h3,h3/l3}
  \draw[flow] (\a) -- (\b);

\coordinate (passgate) at (7.68,4.90);
\coordinate (stopgate) at (4.43,4.90);
\draw[back] (l3.north) -- (11.28,4.90) -- (stopgate);
\draw[back] (8.65,4.90) -- (m3.north);
\draw[back] (5.40,4.90) -- (m2.north);
\draw[hsgdGreen, line width=0.95pt, fill=white] (passgate) circle (0.18cm);
\node[font=\scriptsize\bfseries, text=hsgdGreen] at (passgate) {$\leftarrow$};
\node[tinylabel, text=hsgdGreen] at (7.68,5.28) {$z_3^h=0$};
\node[draw=hsgdRed, fill=white, circle, minimum size=0.38cm,
  inner sep=0pt, font=\scriptsize\bfseries, text=hsgdRed]
  at (stopgate) {$\times$};
\node[tinylabel, text=hsgdRed] at (4.43,5.28) {$z_2^h=1$};
\draw[faintback] (stopgate) -- (3.45,4.90) -- (h1.north);

\draw[hsgdGreen!65, dotted, line width=0.75pt]
  (passgate) -- (7.68,3.92);
\draw[hsgdRed!65, dotted, line width=0.75pt]
  (stopgate) -- (4.43,3.92);
\node[tinylabel, text=hsgdRed] at (3.45,3.20)
  {$\nabla_{\mathbf{h}_t^1}\mathcal L_3=0$};
\node[tinylabel, text=hsgdGreen] at (6.70,3.20)
  {$\nabla_{\mathbf{h}_t^2}\mathcal L_3\neq0$};

\draw[flow] (12.55,4.45) -- (13.35,4.45);
\node[anchor=west, tinylabel, text=hsgdGray] at (13.48,4.45) {forward};
\draw[back] (13.35,3.70) -- (12.55,3.70);
\node[anchor=west, tinylabel, text=hsgdGray] at (13.48,3.70) {gradient};

\fill[hsgdWarm, rounded corners=5pt] (0,0) rectangle (16.4,2.35);
\node[paneltitle] at (0.28,2.08) {(b) Embedding-path detachment};

\node[tinylabel, text=hsgdGreen] at (3.85,1.72) {$z_d^e=0$};
\node[embed] (eleft) at (0.95,1.10) {$E$};
\node[embed] (xleft) at (2.35,1.10) {$\mathbf{e}_{t+d}$};
\node[module] (mleft) at (3.85,1.10) {MTP};
\node[state] (houtleft) at (5.30,1.10) {$\mathbf{h}_t^d$};
\node[loss] (lleft) at (6.62,1.10) {$\mathcal L_d$};
\foreach \a/\b in {eleft/xleft,xleft/mleft,mleft/houtleft,houtleft/lleft}
  \draw[flow] (\a) -- (\b);
\draw[back] (lleft.south) -- (6.62,0.38) -- (0.95,0.38) -- (eleft.south);
\draw[back] (3.85,0.38) -- (mleft.south);

\node[tinylabel, text=hsgdRed] at (12.25,1.72) {$z_d^e=1$};
\node[embed] (eright) at (9.35,1.10) {$E$};
\node[embed] (xright) at (10.75,1.10) {$\mathbf{e}_{t+d}$};
\node[module] (mright) at (12.25,1.10) {MTP};
\node[state] (houtright) at (13.70,1.10) {$\mathbf{h}_t^d$};
\node[loss] (lright) at (15.02,1.10) {$\mathcal L_d$};
\foreach \a/\b in {eright/xright,xright/mright,mright/houtright,houtright/lright}
  \draw[flow] (\a) -- (\b);
\coordinate (estop) at (11.50,0.38);
\draw[back] (lright.south) -- (15.02,0.38) -- (estop);
\draw[back] (12.25,0.38) -- (mright.south);
\draw[faintback] (estop) -- (9.35,0.38) -- (eright.south);
\draw[hsgdRed!65, dotted, line width=0.75pt] (estop) -- (11.50,1.10);
\node[draw=hsgdRed, fill=white, circle, minimum size=0.38cm,
  inner sep=0pt, font=\scriptsize\bfseries, text=hsgdRed] at (estop) {$\times$};

\pgfresetboundingbox
\path[use as bounding box] (-0.05,-0.05) rectangle (16.45,6.05);
\end{tikzpicture}
  \input{figures/hsgd_figure/hsgd_overview_caption.tex}
  \label{fig:hsgd-overview}
\end{figure}

  \subsection{MTP Computation}

  Following Section~\ref{sec:hsm-baseline}, MTP step $d$ combines
  $\mathbf{h}^{d-1}_{t}$ with the shifted token embedding $\mathbf{e}_{t+d}$
  to produce $\mathbf{h}^{d}_{t}$, which predicts $x_{t+d+1}$:
  \[
      \mathbf{h}^{d}_{t}
      =
      M_{\theta_d}
      \left(
          \mathbf{h}^{d-1}_{t},
          \mathbf{e}_{t+d}
      \right),
      \qquad d \in \{1,\ldots,D\}.
  \]
  The parameters $\theta_d$ may be position-specific or shared across all
  stages, as in a recurrent MTP drafter.
  The position-wise notation suppresses dependence on the preceding
  causal context. For the detachment equations below, let
  $\mathbf{u}^{d}_{t}$ denote the actual hidden state input to stage $d$
  before detachment. At the first step,
  $\mathbf{u}^{1}_{t}=\mathbf{h}^{0}_{t}$. At subsequent steps,
  $\mathbf{u}^{d}_{t}=\mathbf{h}^{d-1}_{t}$ under standard MTP, whereas
  with HSM it is the aligned mixed state selected for stage $d$.

  We write $\operatorname{sg}(\cdot)$ for the stop-gradient operator:
  \[
      \operatorname{sg}(\mathbf{x}) = \mathbf{x},
      \qquad
      \frac{\partial\,\operatorname{sg}(\mathbf{x})}
           {\partial \mathbf{x}} = \mathbf{0}.
  \]
  Thus, stop-gradient is the identity during the forward pass and blocks
  the corresponding path only during backpropagation.

  \subsection{Hidden State Detachment}

  For every MTP step after the first, we independently sample a scalar
  Bernoulli variable
  \[
      z^{h}_{d} \sim \operatorname{Bernoulli}(p_h),
      \qquad d \in \{2,\ldots,D\},
  \]
  where $p_h$ is the hidden state detach probability. The hidden state input
  after detachment is then
  \begin{equation}
      \overline{\mathbf{u}}^{d}_{t}
      =
      (1-z^{h}_{d})\,\mathbf{u}^{d}_{t}
      +
      z^{h}_{d}\,\operatorname{sg}
      \left(\mathbf{u}^{d}_{t}\right).
      \label{eq:hidden-state-detach}
  \end{equation}
  The first MTP step always receives the target-model representation
  without detachment, since its input is backbone-produced in both
  training and inference; thus
  $\overline{\mathbf{u}}^{1}_{t}=\mathbf{u}^{1}_{t}=\mathbf{h}^{0}_{t}$.
  When $z^{h}_{d}=1$, losses at stage $d$ and subsequent stages cannot
  propagate through this boundary:
  \[
      \frac{\partial \overline{\mathbf{u}}^{d}_{t}}
           {\partial \mathbf{u}^{d}_{t}}
      =
      (1-z^{h}_{d})\mathbf{I}.
  \]
  Importantly, the parameters used by stage $d$ remain trainable. In the
  shared-parameter setting, a detached stage still contributes gradients
  to the current application of the recurrent MTP module, while gradients
  through earlier unrolled applications along the detached path are
  truncated. HSGD can therefore
  be viewed as stochastic truncation of backpropagation through the MTP
  recurrence rather than removal of the later-position training losses.
  If HSGD is combined with HSM, mixing is performed first to obtain
  $\mathbf{u}^{d}_{t}$, then Equation~\eqref{eq:hidden-state-detach} is
  applied to that input. A detach event consequently blocks gradients through all aligned
  hidden state sources selected for that input, including any selected
  backbone states, while preserving the mixed value in the forward pass.
  Earlier states can receive gradients through other connected
  paths, including their own losses and later HSM selections.

  \subsection{Embedding-Path Detachment}

  As discussed in Section~\ref{sec:hsgd-motivation}, training uses ground-truth token embeddings, whereas later drafting steps consume embeddings of MTP-generated tokens. We regularize this pathway by stochastically detaching the embedding inputs. At each MTP step, we independently sample
  \[
      z^{e}_{d} \sim \operatorname{Bernoulli}(p_e),
      \qquad d \in \{1,\ldots,D\},
  \]
  and replace the embedding input by
  \[
      \overline{\mathbf{e}}_{t+d}
      =
      (1-z^{e}_{d})\,\mathbf{e}_{t+d}
      +
      z^{e}_{d}\,\operatorname{sg}
      \left(\mathbf{e}_{t+d}\right).
  \]
  We apply embedding-path detachment at every MTP step, including the
  first, to regularize the shared embedding-to-drafter pathway. During
  inference, the first MTP step consumes a target-selected token, whereas
  subsequent steps consume draft-generated tokens. A detach event blocks
  the gradient from the associated MTP embedding branch into the shared
  embedding table. It does not freeze the embedding parameters: they
  receive gradients from the target-model language-modeling
  objective and from other connected MTP paths, including the path through
  backbone hidden states. Detachment is applied before the embedding
  normalization and fusion projection, so these components and the current
  MTP module remain trainable.
  The complete stochastic MTP update is therefore:
  \[
      \mathbf{h}^{d}_{t}
      =
      M_{\theta_d}
      \left(
          \overline{\mathbf{u}}^{d}_{t},
          \overline{\mathbf{e}}_{t+d}
      \right).
  \]

  \subsection{Interpretation and Computational Cost}

  HSGD regularizes \emph{credit assignment} rather than the forward
  representation distribution. Because
  $\operatorname{sg}(\mathbf{x})=\mathbf{x}$ in the forward pass, HSGD
  leaves the forward activations and corresponding loss values unchanged
  for fixed parameters and forward inputs. It stochastically suppresses
  selected gradient contributions through upstream hidden states and
  teacher-forced embeddings. This is intended to reduce cross-stage and
  embedding--drafter co-adaptation while preserving direct supervision at
  every recurrent application of the MTP module.

  The method does not introduce additional
  inference-time computation, parameters, or memory. Unless otherwise
  specified, our experiments use $p_h = p_e = 0.3.$
  
  The hidden state and embedding detachment decisions are independent.
  Each eligible MTP stage and microbatch forward pass uses one scalar
  Bernoulli draw per detachment path, shared across all tokens and
  sequences in that microbatch.

\section{Experiments}
\label{sec:experiments}
We first examine design choices for co-training MTP during model pretraining. We then compare HSM and HSGD through pretraining and supervised fine-tuning, examining their effects on acceptance rates across draft positions. Next, we evaluate frozen-backbone \emph{MTP boosting} setting across several models and ablate its main components. Finally, we evaluate HSM as an alternative to TTT for training EAGLE-3 drafters, comparing acceptance quality and training memory.

\subsection{Evaluation Setup and Metrics}
\label{sec:evaluation-setup}

We evaluate speculation quality using vLLM on the \emph{Qualitative} split of
SPEED-Bench~\citep{speedbenchnvidia}, with an output sequence length limit
of 4k tokens. The benchmark aggregates samples from 18 public datasets and
covers 11 semantic domains. Long-input evaluations use the \emph{Throughput}
splits. Unless otherwise specified, we use a draft length (DL) of
seven, with greedy drafter sampling at evaluation temperature $T=0$ and
probabilistic drafter sampling at $T=1$.

We use \emph{acceptance} as a general term for speculation quality. We report two specific metrics: acceptance rates (ARs) and per-sample average acceptance length (AL). For drafted-token position $i$,  $\mathrm{AR}_i$ is the probability that token $i$ is accepted given that all earlier drafted tokens were accepted. AL is the expected number of tokens produced per target-model verification step, including the free bonus token. Mean AL weights each request equally, whereas the reported ARs are pooled globally over all drafted tokens.

\subsection{MTP Co-training Design Choices}
\label{sec:pretraining}
Before evaluating HSM and HSGD, we examine several choices in standard
MTP training during model pretraining: parameter sharing and the number of MTP steps,
whether to jointly optimize the backbone and MTP module, and when to
enable MTP supervision. These experiments motivate the co-training
setup used in the subsequent HSM and HSGD experiments.

  \textbf{Model architecture.}
  We use a hybrid Mamba--Transformer sparse mixture-of-experts
  model following the design principles of Nemotron 3 Nano
  \citep{nvidia2025nemotron3nanoopen}. The model belongs to the approximately
  30B-total/3B-active parameter class, but is trained from scratch for these
  experiments rather than initialized from a released Nemotron checkpoint, unless otherwise stated.

  The experimental backbone contains 54 blocks with model dimension 2,688.
  Its layer pattern comprises 23 Mamba-2 blocks, eight grouped-query
  self-attention blocks, and 23 sparse MoE blocks. The attention blocks use
  32 query heads, eight key--value heads, and a head dimension of 128. The
  Mamba blocks use 64 heads.

Each MoE block contains 128 routed experts, of which six are selected per
token using a learned router with sigmoid gating. Each routed expert has
intermediate dimension 1,856, while the shared-expert pathway has
intermediate dimension 3,712. We use squared-ReLU activations and RMSNorm.
The model has no explicit positional embeddings, no attention or hidden
dropout, and no biases in its linear layers. The input embedding and
LM head weights are not tied. Each MTP module consists of an attention block and an MoE block. The architecture is summarized in
Appendix~\ref{app:pretraining-architecture}.

\textbf{Pretraining configuration.}
The parameter-sharing and MTP supervision timing experiments use the
following training configuration.
We first pretrain each configuration on 1T tokens with a sequence
length of 8k. Training comprises roughly 122M sequences, corresponding to 159k optimization updates. We use a global batch size of 768 sequences.

We train in BF16 using AdamW with $\beta_1=0.9$, $\beta_2=0.95$, weight decay 0.1, and gradient clipping at 1.0. The learning rate follows a Warmup--Stable--Decay schedule. It is linearly increased to $1.2\times10^{-3}$ over roughly 1M sequences, corresponding to approximately 8.39 billion tokens. It remains stable during the middle of training and is decayed over the final 18.3M sequences, corresponding to approximately 150 billion tokens, to a minimum learning rate of $1.2\times10^{-5}$. Every configuration is trained on the same
fixed 1T-token data mixture and with the same tokenizer.

The MoE layers use an auxiliary load-balancing loss with coefficient $10^{-4}$. The router scores are computed in FP32, and expert-bias-based load balancing is enabled. All models use the same initialization, normalization, optimizer, and numerical-precision settings.

The MTP module reuses the backbone's embedding table and language-model head (LM head). We weight the aggregate MTP objective by 0.1 relative to the main next-token objective and distribute  this weight uniformly across all the MTP steps.

\subsubsection{Parameter Sharing and Number of MTP Steps}
\label{sec:mtp-sharing-depth}
Sharing weights across MTP modules is important when the drafter is reused autoregressively at inference. With separate, position-specific heads, each head is trained on inputs from a particular stage of the prediction chain. Sharing parameters instead trains the same module on both target-model representations and representations produced by its earlier applications, aligning training more closely with recurrent drafting. Here, an MTP step is one unrolled application of the MTP module within a training forward pass. We compare MTP trained with position-specific parameters against MTP trained with shared parameters, both using two MTP steps during training, and also experiment with increasing the number of steps to three for the shared-parameter model.

\begin{figure}[t]
\centering
\begin{minipage}[t]{0.42\linewidth}
  \vspace{0pt}
  \begin{minipage}[c][4.8cm][c]{\linewidth}
    \centering
    \includegraphics[width=\linewidth]{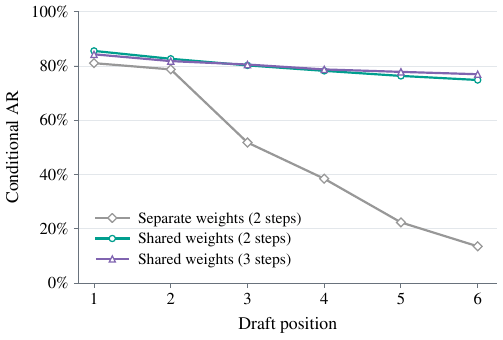}
  \end{minipage}
  \caption{Effect of weight sharing and the number of MTP steps during training on acceptance rates.
  Sharing weights with two MTP steps substantially improves ARs beyond the first
  two positions. Evaluation temperature is $T=0$.%
  }
  \label{fig:mtp-sharing-depth}
\end{minipage}\hfill
\begin{minipage}[t]{0.56\linewidth}
  \vspace{0pt}
  \begin{minipage}[c][4.8cm][c]{\linewidth}
    \centering
    \footnotesize
    \setlength{\tabcolsep}{2pt}
    \renewcommand{\arraystretch}{1.15}
\begin{tabular}{@{}lccrrr@{}}
    \toprule
    Configuration & \shortstack{MTP\\steps} & TFLOP/s & \shortstack{Iteration\\(ms)} &
    \shortstack{Memory\\(GB)}  \\
    \midrule
    Separate & 2 & 659 & 2899 & 140.5 \\
    Shared   & 2 & 648 & 2883 & 139.3 \\
    Separate & 3 & 663 & 3168 & 152.3\\
    Shared & 3   & 662 & 3175 & 150.1\\
    \bottomrule
  \end{tabular}
  \end{minipage}
  \captionof{table}{
  Training performance with separate or shared MTP weights for the Nano-style backbone on 16 NVIDIA GB200 GPUs, using tensor parallelism of 1, expert parallelism of 16, a microbatch size of 2, and 4 gradient accumulation steps. Separate denotes position-specific parameters; Shared denotes parameters shared across prediction positions. Memory reports peak allocated GPU memory.}
  \label{tab:mtp-sharing-performance}
\end{minipage}
\end{figure}

\textbf{Weight sharing preserves acceptance rates at longer draft lengths.} 
At inference, the position-specific baseline applies the first MTP module once and then recurrently reuses the second module for all remaining draft positions. Figure~\ref{fig:mtp-sharing-depth} shows the ARs at the first six draft positions. With two MTP steps during training, weight sharing improves the first-position acceptance rate from $81.04\%$ to $85.55\%$. The benefit is much larger at later positions: the acceptance rate at position three increases from $51.80\%$ to $80.22\%$, and at position six from $13.55\%$ to $74.84\%$, a gain of $61.29$ percentage points. These results support sharing MTP weights when a single head is deployed recurrently, particularly when drafting beyond the number of MTP steps used during training.

\textbf{Two MTP steps offer a favorable training-cost tradeoff.}
Increasing the number of MTP steps during training from two to three with shared parameters changes acceptance rates only marginally (Figure~\ref{fig:mtp-sharing-depth}). The additional training step modestly improves acceptance rates at later draft positions, without improving the first two. Table~\ref{tab:mtp-sharing-performance} reports the corresponding training performance measurements.
The shared-parameter configuration with three MTP steps increases training latency by $10.13\%$.
Peak allocated memory increases from $139.3$ to $150.1$\,GiB ($7.75\%$).
Overall, three MTP steps provide no consistent improvements in acceptance rates. %

At a fixed number of MTP steps, sharing has little effect on iteration time: it decreases by 0.55\% with two steps and increases by 0.22\% with three steps, relative to the corresponding position-specific configuration. Sharing also reduces peak allocated memory from 140.5 to 139.3 GiB (0.85\%) with two steps and from 152.3 to 150.1 GB (1.44\%) with three steps. We therefore use shared MTP with two steps during backbone–MTP co-training.

\subsubsection{Timing of MTP Supervision in Pretraining}
\label{sec:mtp-activation-timing}
Co-training the backbone and MTP allows MTP prediction objectives to shape both the drafter and the target-model representations on which drafting depends. We examine whether these objectives should be enabled from initialization or can be deferred until later in pretraining. Using a fixed 1T-token budget, we find that training MTP from the start yields lower final prediction loss and higher agreement with the target model than every delayed-activation configuration considered.

We vary the update at which MTP prediction supervision begins. For an
activation update $s_{\mathrm{on}}$, the prediction objective is
\begin{equation}
  \mathcal{L}(s)=\mathcal{L}_{\mathrm{LM}}(s)
  +\frac{\lambda_{\mathrm{MTP}}(s)}{2}
    \left[\mathcal{L}_{\mathrm{MTP},1}(s)
          +\mathcal{L}_{\mathrm{MTP},2}(s)\right],
  \qquad
  \lambda_{\mathrm{MTP}}(s)=
  \begin{cases}
    0, & s<s_{\mathrm{on}},\\
    0.1, & s\geq s_{\mathrm{on}}.
  \end{cases}
\end{equation}
We omit the MoE load-balancing term from the objective for clarity. The always-on
configuration uses $s_{\mathrm{on}}=0$. For delayed activation, we resume checkpoints from a single run trained without MTP prediction loss. We use checkpoints at 60k, 90k, 120k, and 140k updates, corresponding to approximately 377B, 566B, 755B, and 881B tokens, respectively. At each checkpoint, we enable MTP prediction loss and continue training to the same 1T-token budget. We preserve the model parameters, optimizer state, and learning-rate schedule, and the backbone remains trainable throughout.

\begin{figure}[!t]
  \centering
  \includegraphics[width=0.65\linewidth,trim=0 187.28775bp 0 0,clip]{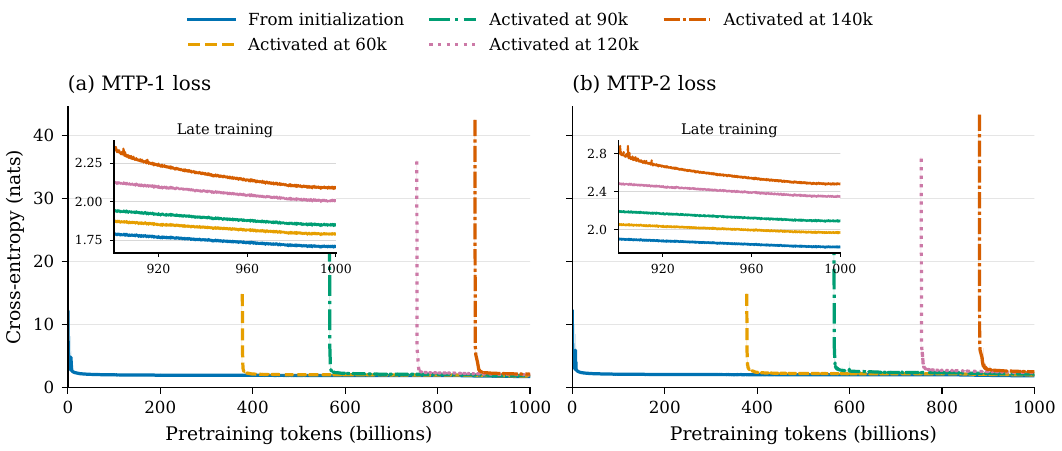}
  \par\vspace{4pt}
  \includegraphics[width=0.49\linewidth,trim=0 0 0 29bp,clip]{figures/mtp_activations/mtp_activation_loss.pdf}\hfill
  \includegraphics[width=0.49\linewidth,trim=0 0 0 29bp,clip]{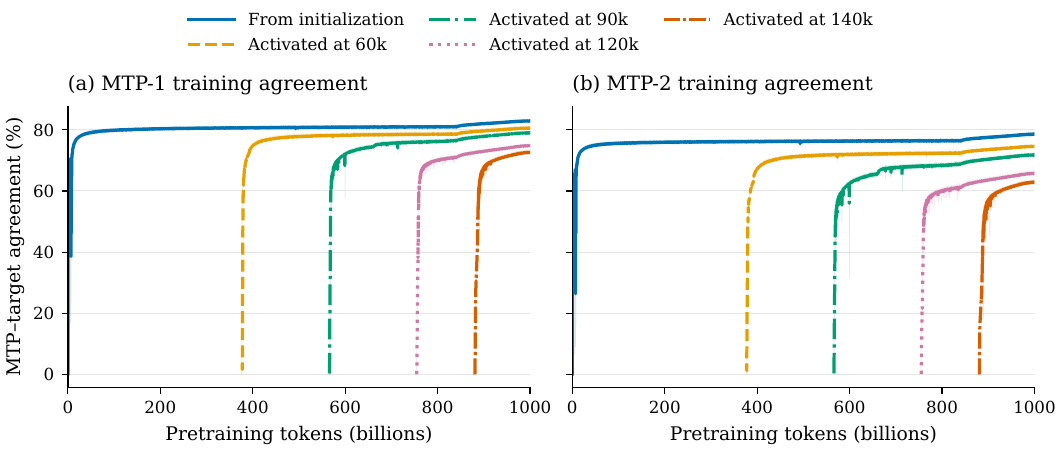}
  \caption{MTP activation timing during 1T-token pretraining.
  Left pair: MTP prediction loss; right pair: teacher-forced MTP-target
  argmax agreement. Each pair shows the first and second shared-MTP steps.
  The horizontal axes measure absolute pretraining tokens; loss insets
  show the final 10\% of the displayed range. Agreement measures alignment
  with target-model predictions, rather than acceptance rates
  during autoregressive drafting.}
  \label{fig:mtp-activation-loss}
  \label{fig:mtp-activation-agreement}
\end{figure}

\textbf{Prediction loss.}
Figure~\ref{fig:mtp-activation-loss} (left pair) shows the training loss at both MTP
steps. Delaying MTP training produces an initial loss spike followed by rapid
adaptation, but the delayed runs remain above the always-on configuration
at the end of the common training budget. The late-training insets reveal
a consistent ordering: longer delays yield higher final loss at both
depths (Table~\ref{tab:mtp-activation-final}).

\textbf{Agreement with the target model.}
Figure~\ref{fig:mtp-activation-agreement} (right pair) reports teacher-forced argmax
agreement between the MTP predictions and the corresponding shifted
target-model predictions. Agreement increases rapidly after activation,
but delayed runs do not reach the always-on configuration by the end of
pretraining. 
Training from initialization reaches final MTP-1
and MTP-2 agreement of 82.89\% and 78.56\%, compared with 80.56\% and
74.60\% for activation at 60k, and 72.63\% and 62.90\% for activation at
140k. Relative to the 140k configuration, supervision from initialization
improves agreement by 10.26 and 15.66 percentage points at the two depths (Table~\ref{tab:mtp-activation-final}).

\textbf{Implications for pretraining.}
Under the fixed 1T-token budget, enabling MTP supervision from initialization improves both prediction loss and target agreement. These results support
enabling MTP prediction supervision from the beginning of pretraining.

\subsubsection{Backbone–MTP Co-training}
\label{sec:backbone_mtp_cotraining}
We investigate the impact of jointly optimizing the backbone and MTP module during continued training on pretraining data. We initialize the student from pretrained Nemotron 3 Nano and use pretrained Nemotron 3 Super~\citep{nvidia2026nemotron3superopen} as the distillation teacher. We compare two configurations trained for 100B tokens each: \emph{frozen-backbone training}, which updates only the MTP module, and \emph{co-training}, which updates both the backbone and MTP parameters.

\begin{figure}[t]
\centering
\begin{minipage}[t]{0.49\linewidth}
\vspace{0pt}
\centering
\begin{minipage}[c][4.2cm][c]{\linewidth}
\centering
\small
\setlength{\tabcolsep}{3pt}
\begin{tabular*}{\linewidth}{@{\extracolsep{\fill}}lrrrr@{}}
\toprule
 & \multicolumn{2}{c}{Loss $\downarrow$}
 & \multicolumn{2}{c}{Agreement (\%) $\uparrow$} \\
\cmidrule(lr){2-3}\cmidrule(lr){4-5}
Activation & MTP-1 & MTP-2 & MTP-1 & MTP-2 \\
\midrule
Initialization & \textbf{1.70} & \textbf{1.80} & \textbf{82.89} & \textbf{78.56} \\
60k updates & 1.78 & 1.95 & 80.56 & 74.60 \\
90k updates & 1.84 & 2.08 & 79.01 & 71.82 \\
120k updates & 1.99 & 2.33 & 74.87 & 65.81 \\
140k updates & 2.08 & 2.47 & 72.63 & 62.90 \\
\bottomrule
\end{tabular*}
\end{minipage}
\captionof{table}{Final logged training metrics after approximately 1T tokens. Prediction losses are cross-entropy; target agreement is reported in percent.}
\label{tab:mtp-activation-final}
\end{minipage}\hfill
\begin{minipage}[t]{0.48\linewidth}
\vspace{0pt}
\centering
\begin{minipage}[c][4.2cm][c]{\linewidth}
\centering
\includegraphics[width=\linewidth]{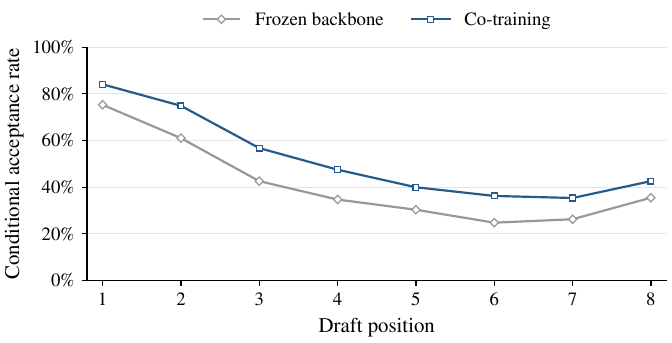}
\end{minipage}
\caption{Impact of backbone--MTP co-training on acceptance rates
at draft positions 1--8 for Nemotron 3 Nano. Co-training improves acceptance rates at all eight displayed positions. }
\label{fig:backbone_mtp_cotraining}
\end{minipage} 
\end{figure}

Figure~\ref{fig:backbone_mtp_cotraining} reports acceptance rates at the first eight draft positions. Co-training improves acceptance rates at every position shown. At the first three positions, acceptance rates increase from 75.30\%, 61.03\%, and 42.59\% to 84.09\%, 74.88\%, and 56.76\%, respectively.
These results highlight the benefit of allowing the target model
to adapt alongside its drafter. Joint optimization enables backbone representations to evolve in support of MTP prediction, providing a mechanism for improving backbone--drafter alignment. %

\subsection{HSM and HSGD through Pretraining and Supervised Fine-Tuning (SFT)}
  \label{sec:pretraining-sft-experiments}

This section compares standard shared-parameter MTP, HSM, HSGD, and their combination through pretraining and supervised fine-tuning, and discusses how their different AR profiles may benefit inference at different serving concurrencies.

\subsubsection{Experimental Setup}
We study \hsm and HSGD in a co-training regime in which the target model
and its MTP drafter are trained jointly from initialization. In this setting, the auxiliary MTP
loss can update the target backbone, shared token embeddings, output
projection, and MTP parameters.

We follow the architecture and pretraining setup in
Section~\ref{sec:pretraining}, with shared MTP parameters and two
MTP steps during training. We evaluate four configurations
to isolate the individual and combined effects of \hsm and HSGD: 
We compare the shared-MTP baseline with variants
using \hsm, HSGD, or both. When HSGD is enabled, we set the
hidden state and embedding-path detach probabilities to
$p_h=p_e=0.3$.
All other architectural, data, optimization, and training
settings are held fixed across configurations. When enabled,
\hsm{} and HSGD remain active throughout short-context
pretraining, long-context continuation, and supervised
fine-tuning.

\textbf{Short-context pretraining.}
All configurations use the pretraining setup described in
Section~\ref{sec:pretraining}.

\textbf{Long-context continuation.}
Following short-context pretraining, we continue training each configuration with a sequence length of 128k tokens. This stage uses a global batch size of 128 sequences, corresponding to approximately 16.78M tokens per update. We use a constant learning rate of $4.5\times10^{-6}$ and retain the AdamW hyperparameters, weight decay, gradient clipping, BF16 precision, MoE configuration, and MTP loss weight used during short-context pretraining. Long-context continuation uses eight-way context parallelism and the same long-context data mixture for every configuration. 
For the controlled comparison, we select the checkpoint after 4,000 long-context updates. At the configured sequence length and global batch size, this checkpoint has processed approximately 67.1B tokens during long-context continuation.

\textbf{SFT configuration.}
Starting from the long-context pretrained checkpoints, we apply the same SFT procedure to all four configurations.
We perform SFT with sequence packing to a maximum sequence length of 128k tokens and a global batch size of 256 packed sequences. %
All configurations use the same packed SFT data mixture, tokenizer, prompt format, and sample ordering.
We optimize in BF16 using AdamW with $\beta_1=0.9$, $\beta_2=0.95$, weight decay 0.1, and gradient clipping at 1.0. The learning rate is warmed up over 25,600 samples, equivalent to 100 optimization updates, to a peak value of $10^{-5}$. It subsequently follows a cosine schedule with a specified minimum learning rate of $2\times10^{-6}$ and a decay horizon of roughly 2M samples. The MTP loss scaling factor remains 0.1.

\subsubsection{Acceptance Results and Tradeoffs}
We report acceptance rates at a draft length of 11 and evaluate at $T=0$. Figure~\ref{fig:results-overview}(a) compares the acceptance rate curves for shared-parameter MTP, HSM, HSGD, and their combination. Table~\ref{tab:pretraining-sft-mean-al} reports their mean ALs.
HSGD achieves the highest acceptance rates at the first four draft
positions, whereas \hsm achieves the highest ARs at positions five
through eleven. At draft length 11, \hsm yields
the highest mean AL of 5.57, compared with 5.24 for HSGD and 5.19 with shared MTP.

These acceptance rate curves suggest that the preferred training method depends on the \emph{serving target}. Higher-concurrency serving often favors shorter draft lengths to limit the additional verification cost, making high acceptance rates at the first few positions particularly valuable. HSGD's stronger early-position ARs may therefore translate into better end-to-end speedups in this regime. At lower concurrency, decoding is often memory-bandwidth-bound, and longer drafts can amortize target-model weight reads across more tokens. This regime favors \hsm's higher tail acceptance rates. Combining \hsm and HSGD provides no additional benefit over \hsm alone in this setting.

\begin{table}[htbp]
  \centering
  \small
  \setlength{\tabcolsep}{4pt}
  \renewcommand{\arraystretch}{1.35}
  \begin{tabular}{@{}lr@{}}
\toprule
Method & AL \\
\midrule
Shared-parameter MTP & 5.19 \\
\hsm & \textbf{5.57} \\
\hsgd & 5.24 \\
\hsm+\hsgd & 5.47 \\
\bottomrule
\end{tabular}
  \caption{Mean AL after pretraining and SFT at $T=0$ and draft length
  11. Bold indicates the highest AL. Per-domain results appear in
  Appendix~\ref{app:pretraining-sft-results}.}
  \label{tab:pretraining-sft-mean-al}
\end{table}

\subsubsection{Downstream Performance After Pretraining}

We also evaluate downstream model quality at the 1T-token
pretraining checkpoint, before long-context continuation and SFT.
The shared-parameter MTP baseline and its HSM and HSGD variants achieve
comparable mean scores across 17 common evaluation metrics, suggesting that HSM and HSGD do not compromise downstream model quality. Full results appear in
Appendix~\ref{app:hsm_downstream}.

\subsection{Frozen-Backbone \mtp Boosting} 
\label{sec:frozen-backbone-mtp-training}
We use \emph{MTP boosting} to denote continued training, i.e. finetuning, of an existing
MTP module with all target/backbone-model parameters frozen, including the shared
token embeddings and LM head. In this setting,
we use logit-level knowledge distillation from backbone logits.

\subsubsection{MTP Boosting with HSM}
\label{sec:hsm-boosting}
We evaluate the \emph{MTP boosting} procedure using HSM on four Nemotron models: \super{}~\citep{nvidia2026nemotron3superopen}, \super3dot5{}~\citep{nvidia2026nemotron35},
\lightning{}~\citep{nvidia2026nemotron35} and \ultra{}~\citep{nvidia2026nemotron3ultra}. We additionally boosted DeepSeek-V4-Pro's MTP~\citep{deepseekai2026deepseekv4}.
For \super{} and DeepSeek-V4-Pro, MTP boosting was applied on the official released checkpoint. For \super{}, the updated
MTP module was released separately as MTPv2\footnote{\url{https://huggingface.co/nvidia/Nemotron-3-Super-120B-A12B-BF16-MTPv2}}.
For the other three Nemotrons, boosting was part of the official training pipeline. We additionally study adaptation of \super{}'s MTP module to Nemotron-Labs-3-Puzzle-75B-A9B~\citep{bercovich2026nemotronlabs3puzzle}, a compressed and distilled
derivative.

For training data, seed prompts are drawn from
\textit{Nemotron-Post-Training-Dataset-v2}\footnote{\url{https://huggingface.co/datasets/nvidia/Nemotron-Post-Training-Dataset-v2}}
(V2) and \textit{Nemotron-RL-Super-Training-Blends}\footnote{\url{https://huggingface.co/datasets/nvidia/Nemotron-RL-Super-Training-Blends}}
(V3), covering general-purpose and agentic inputs, respectively.
MTP boosting uses forward KL distillation for \super{} and \ultra{}, and LK distillation~\citep{samarin2026lk} for \super3dot5{}, \lightning{}, and DeepSeek-V4-Pro.
Model-specific training configuration and data construction are detailed
in Appendix~\ref{app:boosting-training}.

\textbf{Acceptance length.}
Table~\ref{tab:mtp-boosting-summary} reports mean AL on the SPEED-Bench
qualitative split with draft length seven, following
Section~\ref{sec:evaluation-setup}. 
Boosting improves mean AL for all five models at both evaluation
temperatures, with relative gains ranging from 4.3\% to 51.5\%
at $T=0$ and from 10.8\% to 52.2\% at $T=1$.
DeepSeek-V4-Pro shows the largest improvement, increasing from
2.77 to 4.20 and from 2.66 to 4.05, respectively.
Improvements extend to all 11 domains
(Appendix~\ref{app:boosting-domains}).

\begin{table}[htbp]
  \centering
  \small
  \begin{tabular}{lrrrr}
    \toprule
    & \multicolumn{2}{c}{$T=0$} & \multicolumn{2}{c}{$T=1$} \\
    \cmidrule(lr){2-3}\cmidrule(lr){4-5}
    Model & Pre boosting & \textbf{Boosted} & Pre boosting & \textbf{Boosted} \\
    \midrule
    \super{}       & 3.45 & \textbf{4.32} &  3.10 & \textbf{4.20} \\
    \super3dot5{}     & 4.29 & \textbf{4.72} & 3.79 & \textbf{4.61} \\
    \lightning{} & 3.48 & \textbf{3.85} & 3.13 & \textbf{3.70} \\
    \ultra{}       & 4.39 & \textbf{4.58} & 4.10 & \textbf{4.55} \\
    DeepSeek-V4-Pro & 2.77 & \textbf{4.20} & 2.66 & \textbf{4.05} \\
    \bottomrule
  \end{tabular}
  \caption{Mean AL before and after \mtp boosting with HSM on the
  SPEED-Bench qualitative split with draft length seven.}
  \label{tab:mtp-boosting-summary}
\end{table}

\begin{figure}[!htb]
  \centering
  \includegraphics[width=\linewidth]{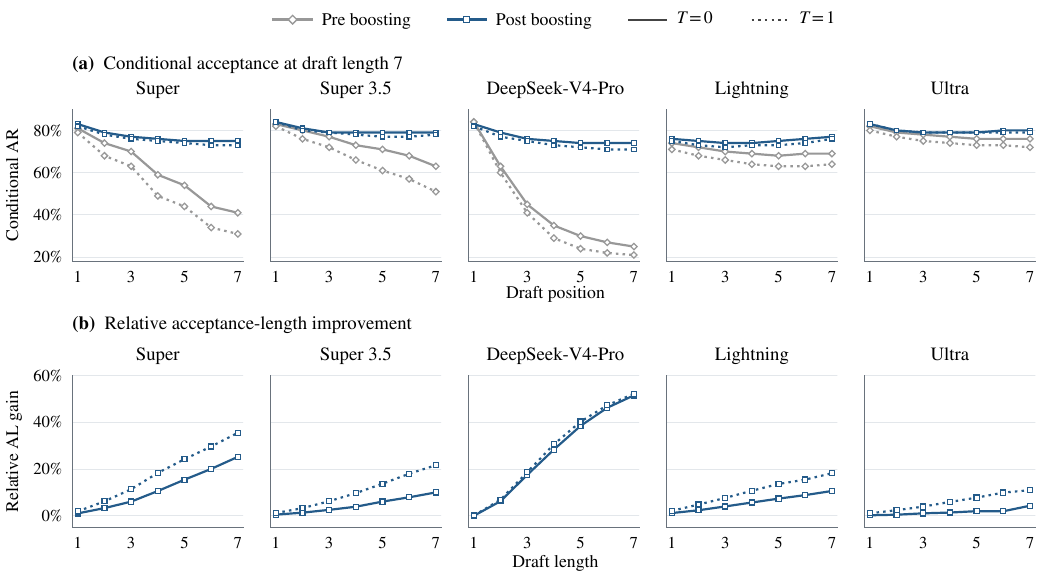}
  \caption{Effect of MTP boosting on the tested models. (a) Acceptance rates
  at draft length seven. (b) Relative AL improvement
  across draft lengths. Solid and dotted curves denote
  greedy decoding at $T=0$ and probabilistic drafting at $T=1$,
  respectively. Boosting improves or preserves AL at every evaluated DL
  for all five models and both temperatures. Full ALs are reported in
  Appendix~\ref{app:boosting-al-by-draft-length}.}
  \label{fig:mtp-boosting-ar}
\end{figure}

\textbf{Acceptance rates by draft position.}
Figure~\ref{fig:mtp-boosting-ar}(a) shows acceptance rates by
draft-token position. The numerical results are reported in Appendix~\ref{app:boosting-ar}.
For the four Nemotron models, MTP boosting maintains or improves acceptance rates
at every reported position, with larger gains at later positions. At $T=0$, \super{}'s first-token acceptance rate rises
from 0.81 to 0.83, while seventh-token acceptance rate rises from 0.41 to
0.75. \super3dot5{} exhibits a similar pattern: seventh-token acceptance rate
increases from 0.63 to 0.79 at $T=0$ and from 0.51 to 0.78 at $T=1$.
\lightning{} and \ultra{} start from flatter AR curves yet still achieve higher
acceptance rates after boosting.
DeepSeek-V4-Pro shows the most substantial gains at positions 2--7.

\textbf{Acceptance length by draft length.}
Figure~\ref{fig:results-overview}(b) summarizes AL gains at draft length seven. Figure~\ref{fig:mtp-boosting-ar}(b) reports the relative AL improvement
at DL 1--7. Boosting improves or preserves AL at every DL for all five models and both evaluation
temperatures. The gains are generally larger at longer draft lengths. Appendix~\ref{app:boosting-al-by-draft-length} reports full results. 

\textbf{Variation across models.}
Models with sharper declines in ARs across draft positions tend to show larger gains from boosting. Gains are particularly large for \super{} and DeepSeek-V4-Pro, whose acceptance rates decline sharply across draft positions. \lightning{} and \ultra{} start from flatter acceptance rate curves, showing smaller improvement. This pattern is consistent with \hsm targeting training--inference mismatch in recursive drafting. Differences in MTP co-training may also contribute to the variation in gains. For \lightning{}, MTP was not co-trained from scratch and for \super{} MTP was initially trained position-specific.
We hypothesize that multi-step boosting helps compensate for
DeepSeek-V4-Pro's single-step MTP pretraining~\citep{deepseekai2026deepseekv4},
contributing to its larger gains.

\textbf{Adaptation to a Compressed Backbone.}
We study MTP boosting with HSM on Nemotron-Labs-3-Puzzle-75B-A9B, a compressed and distilled derivative of \super{}
in Appendix~\ref{app:puzzle-adaptation}.

\subsubsection{MTP Boosting with HSGD}
\begin{figure}[!h]
\centering
\begin{minipage}[t]{0.49\linewidth}
  \vspace{0pt}
  \begin{minipage}[c][6.3cm][c]{\linewidth}
    \centering
    \includegraphics[width=\linewidth]{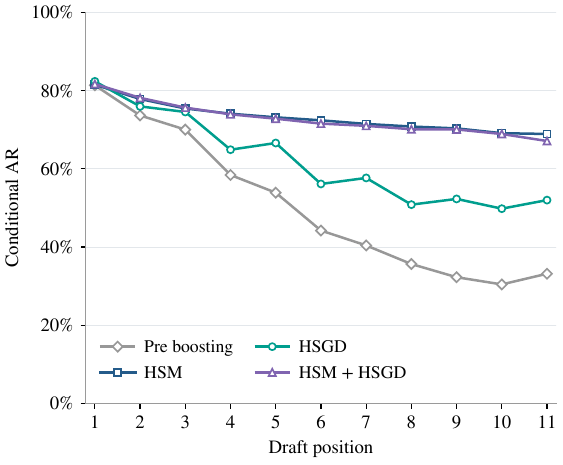}
  \end{minipage}
  \caption{Acceptance rates for Nemotron 3 Super before and after MTP boosting
  on the SPEED-Bench qualitative split with draft length 11.
  Exact values appear in Appendix~\ref{app:hsm-hsgd-boosting-ar}.}
  \label{fig:hsm-hsgd-boosting-ar}
\end{minipage}\hfill
\begin{minipage}[t]{0.49\linewidth}
  \vspace{0pt}
  \begin{minipage}[c][6.3cm][c]{\linewidth}
    \centering
    \small
    \setlength{\tabcolsep}{3pt}
\begin{tabular}{@{}lrrrr@{}}
      \toprule
      \textbf{Domain} & \multicolumn{1}{c}{\textbf{Pre}}
        & \multicolumn{1}{c}{\textbf{HSM}}
        & \multicolumn{1}{c}{\textbf{HSGD}}
        & \multicolumn{1}{c@{}}{\makecell[c]{\textbf{HSM +}\\\textbf{HSGD}}} \\
      \midrule
      Coding        & 3.75 & \textbf{5.04} & 4.28 & 4.93 \\
      Humanities    & 3.25 & \textbf{4.08} & 3.66 & 4.07 \\
      Math          & 3.75 & \textbf{4.98} & 4.18 & 4.97 \\
      Multilingual  & 4.03 & 5.61 & 4.43 & \textbf{5.64} \\
      QA            & 3.20 & \textbf{3.93} & 3.59 & 3.90 \\
      RAG           & 3.85 & 4.99 & 4.19 & \textbf{5.02} \\
      Reasoning     & 3.66 & \textbf{4.66} & 4.04 & 4.62 \\
      Roleplay      & 2.85 & \textbf{3.40} & 3.09 & 3.38 \\
      STEM          & 3.32 & \textbf{4.25} & 3.67 & 4.19 \\
      Summarization & 3.63 & \textbf{4.40} & 3.86 & 4.29 \\
      Writing       & 3.04 & 3.64 & 3.37 & \textbf{3.72} \\
      \midrule
      \textbf{Mean AL} & 3.49 & \textbf{4.45} & 3.85 & 4.43 \\
      \bottomrule
    \end{tabular}
  \end{minipage}
  \captionof{table}{Acceptance length (AL) for Nemotron 3 Super before and after MTP boosting on the SPEED-Bench qualitative split with draft length 11. Pre denotes the pre-boosting checkpoint.}
  \label{tab:mtp-boosting-al}
\end{minipage}
\end{figure}

We compare HSM, HSGD, and their combination on Nemotron 3 Super,
using the same boosting setup as in
Section~\ref{sec:hsm-boosting}. We evaluate all variants at
draft length 11 and evaluation temperature $T=0$.
As shown in Figure~\ref{fig:hsm-hsgd-boosting-ar} and Table~\ref{tab:mtp-boosting-al}, HSM substantially improves acceptance rates at later draft positions, flattening the declining acceptance rates of the original drafter. 
Boosting with HSGD alone also improves both acceptance rates and acceptance length, increasing mean AL to $3.85$. However, its acceptance rates remain below those of HSM throughout positions two through eleven. This improvement may largely reflect continued knowledge distillation: KD-only boosting also achieves a similar AL, as shown in Section~\ref{sec:ablations}.

Combining HSGD with HSM yields slightly higher acceptance rates at the second and third positions, but slightly lower rates than HSM alone throughout the remaining positions. Its mean AL of $4.43$ also remains below that of HSM alone. While HSGD has shown gains in a co-training setting, the results here do not establish an additional benefit from HSGD in an MTP boosting setup.
Moreover, freezing the shared embeddings removes the effect of embedding-path detachment. We hypothesize that this partly explains why applying HSGD doesn't provide strong benefits in this setting.

\subsection{MTP Boosting Ablation Experiments}
\label{sec:ablations}

We ablate \hsm and its mixing strategy, the training objective and data, and the number of \mtp steps during training in an MTP boosting setting and report the effect of each component on draft quality. We additionally report long-context evaluations and measure whether the acceptance length gains persist beyond the training sequence length.

\textbf{Experimental setup.}
We keep the target model frozen as in
Section~\ref{sec:frozen-backbone-mtp-training}.
Unless otherwise specified, we compared results after 10k training iterations, and follow the experiment configuration described in Appendix~\ref{app:boosting-data}.
We conduct all ablations on \super{} and additionally evaluate loss functions, training data, and
the number of MTP steps on \lightning{}. Settings that differ from this configuration are specified in the corresponding experiments. %

\begin{figure}[!htb]
  \centering
  \begin{minipage}[t]{0.64\linewidth}
    \begin{minipage}[c][175pt][c]{\linewidth}
      \centering
      \includegraphics[width=\linewidth]{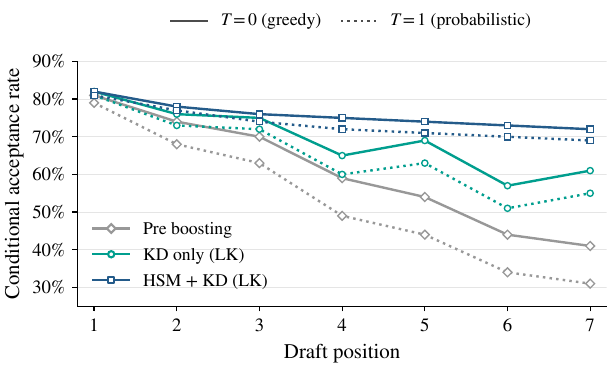}
    \end{minipage}
    \caption{Acceptance rates for \super{} after 10k iterations of KD-only or HSM+KD training with LK.}
    \label{fig:super-hsm-ablation-ar}
  \end{minipage}\hfill
  \begin{minipage}[t]{0.34\linewidth}
    \begin{minipage}[c][175pt][c]{\linewidth}
      \centering
      \small
      \setlength{\tabcolsep}{4pt}
      \renewcommand{\arraystretch}{1.35}
      \begin{tabular*}{\linewidth}{@{\extracolsep{\fill}}lr@{}}
        \toprule
        Method & AL \\
        \midrule
        \multicolumn{2}{c}{$T=0$ (greedy)} \\
        \addlinespace[2pt]
        KD only (LK) & 3.83 \\
        HSM+KD (LK) & \textbf{4.22} \\
        \midrule
        \multicolumn{2}{c}{$T=1$ (probabilistic)} \\
        \addlinespace[2pt]
        KD only (LK) & 3.58 \\
        HSM+KD (LK) & \textbf{4.02} \\
        \bottomrule
      \end{tabular*}
    \end{minipage}
    \captionof{table}{KD vs HSM effect on AL after 10k MTP boosting iterations.}
    \label{tab:super-hsm-al-compact}
  \end{minipage}
\end{figure}

\textbf{Effect of \hsm.}
Applying \hsm on top of KD improves AL over KD-only
training by 10.1\% at $T=0$ and 12.2\% at $T=1$
(Table~\ref{tab:super-hsm-al-compact}).
The first-draft-token acceptance rate remains similar, while acceptance rates
improve at every subsequent position, with generally larger gains
at later positions and at $T=1$.
Figure~\ref{fig:super-hsm-ablation-ar} shows that \hsm produces a
flatter acceptance rate curve, reducing the pronounced drops seen
with KD alone.
Full ARs appear in
Appendix~\ref{app:hsm-acceptance-rates}.

\begin{figure}[!htb]
  \centering
  \begin{minipage}[t]{0.695\linewidth}
    \vspace{0pt}
    \begin{minipage}[c][4.2cm][c]{\linewidth}
      \centering
      \includegraphics[width=\linewidth]{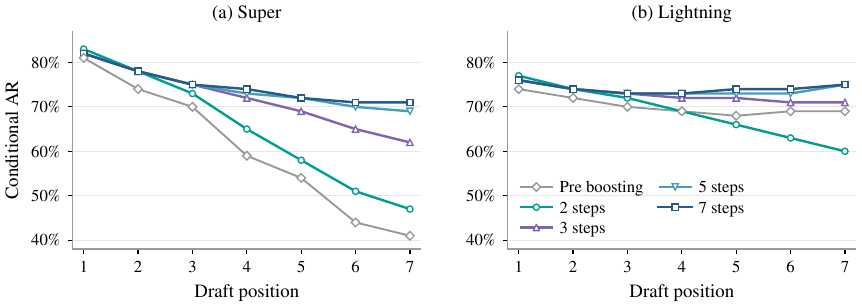}
    \end{minipage}
    \caption{Effect of the number of MTP steps during training with \hsm on acceptance rates at
    $T=0$ and draft length seven: (a) \super{} and (b) \lightning{}. Full results appear in
    Appendix~\ref{app:recurrence-depth-t0}.}
    \label{fig:mtp-steps-t0-conditional-ar}
  \end{minipage}\hfill
  \begin{minipage}[t]{0.285\linewidth}
    \vspace{0pt}
    \begin{minipage}[c][4.2cm][c]{\linewidth}
      \centering
      \small
      \setlength{\tabcolsep}{3pt}
      \renewcommand{\arraystretch}{1.35}
      \begin{tabular}{@{}cccc@{}}
\toprule
\multicolumn{1}{c}{\makecell[c]{Steps}} &
\multicolumn{1}{c}{\makecell[c]{Super}} &
\multicolumn{1}{c}{\makecell[c]{Lightning}} &
\multicolumn{1}{c}{\makecell[c]{Super\\3.5}} \\
\midrule
Pre & 3.45 & 3.48 & 4.29 \\
2 & 3.78 & 3.59 & 4.30 \\
3 & 4.04 & 3.71 & 4.54 \\
5 & 4.12 & 3.75 & 4.60 \\
7 & \textbf{4.14} & \textbf{3.77} & \textbf{4.64} \\
\bottomrule
\end{tabular}
    \end{minipage}
    \captionof{table}{AL by MTP training steps at $T=0$, DL seven. Pre denotes pre-boosting; bold marks the highest AL per model.}
    \label{tab:mtp-steps-mean-al}
  \end{minipage}
\end{figure}

\textbf{Number of MTP steps during training.}
We vary the number of training \mtp steps with a fixed draft length of seven. Increasing the number of steps improves AL
on \super{} and \lightning and Super, with the largest gain from two
to three steps and diminishing returns thereafter
(Table~\ref{tab:mtp-steps-mean-al}).
The gains are concentrated at later draft positions
(Figure~\ref{fig:mtp-steps-t0-conditional-ar}). AR@7 increases from 0.47 to 0.71
on \super{} and from 0.60 to 0.75 on \lightning{}, while AR@1
remains consistent. The gain from two to three steps contrasts with the pretraining results~(Section~\ref{sec:mtp-sharing-depth}), where increasing MTP 
training steps yields marginal changes. 
Full results appear in
Appendix~\ref{app:recurrence-depth-t0}, with additional
\lightning{} and \super3dot5{} results at $T=1$ in
Appendix~\ref{app:recurrence-depth-t1}.
Appendix~\ref{app:mtp-training-cost} reports the
training-time and peak-memory overheads.

\textbf{Loss Terms.}
We compare training with CE on ground-truth data to KL and LK distillation (Table~\ref{tab:loss_ablation}). Both distillation objectives outperform CE on both models, with small AL gains at $T=0$ (at most 0.04 tokens) and larger gains at $T=1$ (5.2--9.7\%). This pattern is consistent with the evaluation criteria:
greedy verification requires agreement with the target's argmax,
whereas rejection sampling depends on the overlap between the
draft and target distributions.
Appendix~\ref{app:kl-loss-temperature} examines the
effect of KL loss temperature.

\begin{table}[!htb]
\centering
\small
\setlength{\tabcolsep}{7pt}
\renewcommand{\arraystretch}{1.12}
\begin{tabular}{@{}l rr rr@{}}
\hline
& \multicolumn{2}{c}{\super{}} & \multicolumn{2}{c}{\lightning{}} \\
\cline{2-3}\cline{4-5}
Loss & $T=0$ &  $T=1$ &  $T=0$ & $T=1$ \\
\hline
CE & 4.19 & 3.82 & 3.72 & 3.30 \\
LK & 4.22 & 4.02 & \textbf{3.76} & 3.56 \\
KL & \textbf{4.23} & \textbf{4.09} & 3.74 & \textbf{3.62} \\
\hline
\end{tabular}
\caption{Effect of the training objective on AL for \super{}
and \lightning{}. Bold values indicate
the highest AL for each model and evaluation temperature.}
\label{tab:loss_ablation}
\end{table}

\noindent\textbf{On-policy Rollouts versus off-policy responses.}
On \super{} and V2 data, we compare our default on-policy setup, which uses backbone-generated responses at $\tau=1$, with an off-policy variant that retains the original dataset responses. On-policy training yields higher AL at both evaluation temperatures
(Table~\ref{tab:rollout_response_source}).

\begin{table}[!htp]
\centering
\small
\setlength{\tabcolsep}{6pt}
\renewcommand{\arraystretch}{1.12}
\begin{tabular}{@{}l r r@{}}
\hline
Response source & $T=0$ & $T=1$ \\
\hline
Off-policy & 3.99 & 3.81 \\
On-policy & \textbf{4.23} & \textbf{4.04} \\
\hline
\end{tabular}
\caption{Effect of response source on \super{} ALs with V2
training data.}
\label{tab:rollout_response_source}
\end{table}

\begingroup
\newpage
\noindent\textbf{Rollout generation temperature.}
We compare on-policy responses generated greedily ($\tau=0$) and through
stochastic sampling ($\tau=1$) for \super{} using V2+V3 prompts. 
Table~\ref{tab:rollout_generation_temperature} shows that AL
differs by less than 1\% between generation temperatures, with no consistent
advantage to either setting. These results indicate limited sensitivity to rollout generation temperature. 

\begin{table}[!htp]
\centering
\small
\setlength{\tabcolsep}{5pt}
\renewcommand{\arraystretch}{1.12}
\begin{tabular}{@{}l c r r@{}}
\hline
Loss & Gen. $\tau$ & $T=0$ & $T=1$ \\
\hline
LK & 0 & \textbf{4.23} & \textbf{4.03} \\
LK & 1 & 4.22 & 4.02 \\
\hline
KL & 0 & \textbf{4.24} & 4.06 \\
KL & 1 & 4.23 & \textbf{4.09} \\
\hline
\end{tabular}
\caption{Mean AL vs rollout generation temperature on \super{}. }
\label{tab:rollout_generation_temperature}
\end{table}

\par
\endgroup

\noindent\textbf{Data mixture.}
We compare training the \super{} and \lightning{} models on V2, V3, and V2+V3 data (Table~\ref{tab:super_rollout_data_ablation}).
On the SPEED-Bench qualitative split, V2 and V2+V3 yield similar ALs, while V3 alone yields slightly lower values. The differences are
small in these experiments, indicating limited sensitivity to the choice
of data source. %

\begin{table}[!h]
\centering
\small
\setlength{\tabcolsep}{3pt}
\renewcommand{\arraystretch}{1.12}
\begin{tabular}{@{}l r r@{}}
\hline
Training data & $T=0$ & $T=1$ \\
\hline
\multicolumn{3}{c}{\super{}} \\
\hline
V2 on-policy & 4.23 & 4.04 \\
V3 on-policy & 4.17 & 3.99 \\
V2+V3 on-policy & 4.22 & 4.02 \\
\hline
\multicolumn{3}{c}{\lightning{}} \\
\hline
V2 on-policy & 3.76 & 3.56 \\
V3 on-policy & 3.73 & 3.54 \\
V2+V3 on-policy & 3.76 & 3.56 \\
\hline
\end{tabular}
\caption{Effect of training data on Mean AL for \super{} and
\lightning{}.}
\label{tab:super_rollout_data_ablation}
\end{table}

\textbf{Generalization to longer inputs.}
We assess whether the gains from \mtp boosting, performed with a max 8k
context length, carry over to longer inputs using the SPEED-Bench throughput
splits, which span from 1k to 32k tokens
(Table~\ref{tab:super_long_isl}). The 1k, 2k, and 8k splits fall within the
boosting context length and serve as a reference, whereas the 16k and 32k
splits extend beyond that length. MTP boosting improves AL by 24\% to 38\% across context lengths. The relative improvement remains consistent and does not degrade beyond the boosting context length. 

\begin{table}[!htp]
\centering
\small
\setlength{\tabcolsep}{4pt}
\renewcommand{\arraystretch}{1.12}
\begin{tabular}{@{}c l r r r r r@{}}
\hline
$T$ & Checkpoint & 1k & 2k & 8k & 16k & 32k \\
\hline
0 & Pre boosting & 3.39 & 3.41 & 3.51 & 3.55 & 3.56 \\
0 & Boosted & \textbf{4.24} & \textbf{4.24} & \textbf{4.42} & \textbf{4.55} & \textbf{4.64}\\
\hline
1 & Pre boosting & 3.03 & 3.05 & 3.11 & 3.15 & 3.19 \\
1 & Boosted & \textbf{4.07} & \textbf{4.12} & \textbf{4.26} & \textbf{4.31} & \textbf{4.43} \\
\hline
\end{tabular}
\caption{ALs before and after MTP boosting for \super{} across different context lengths.}
\label{tab:super_long_isl}
\end{table}

\FloatBarrier

\textbf{HSM Mixing Strategy.}
We compare uniform layer sampling with exponential sparse replacement
(ESR), defined in Section~\ref{sec:hsm-mixing-strategies}. Uniform layer sampling achieves slightly higher ALs (Table~\ref{tab:super_hsm_sampling_ablation}).

\begin{table}[!htp]
\centering
\small
\setlength{\tabcolsep}{6pt}
\renewcommand{\arraystretch}{1.12}
\begin{tabular}{@{}l r r@{}}
\hline
HSM strategy & $T=0$ & $T=1$ \\
\hline
Uniform layer sampling & \textbf{4.22} & \textbf{4.02} \\
ESR & 4.20 & 3.99 \\
\hline
\end{tabular}
\caption{Effect of the HSM mixing strategy in \super{} MTP boosting.}
\label{tab:super_hsm_sampling_ablation}
\end{table}

\FloatBarrier

\subsection{\hsm as an EAGLE-3/\ttt Alternative}
\label{sec:eagle3-alternative}

We evaluate \hsm as an alternative to EAGLE-3's training-time test
(\ttt), described in Section~\ref{sec:ttt-cost}, comparing acceptance rates, acceptance lengths, and training cost.

\begin{table}[t]
  \centering
  \small
    \setlength{\tabcolsep}{1.5pt}
  \resizebox{\textwidth}{!}{%
\begin{tabular}{@{}l*{4}{c}|*{4}{c}|*{4}{c}|*{3}{c}|*{3}{c}@{}}
    \toprule
    & \multicolumn{4}{c|}{\shortstack{\textbf{Llama-3.1-}\\\textbf{8B-Instruct}}}
    & \multicolumn{4}{c|}{\textbf{Qwen3-8B}}
    & \multicolumn{4}{c|}{\shortstack{\textbf{Qwen3.5-}\\\textbf{122B-A10B}}}
    & \multicolumn{3}{c|}{\shortstack{\textbf{GPT-OSS-}\\\textbf{120B}}}
    & \multicolumn{3}{c}{\shortstack{\textbf{Qwen3-235B-}\\\textbf{Instruct}}} \\
    \cmidrule(lr){2-5}\cmidrule(lr){6-9}\cmidrule(lr){10-13}\cmidrule(lr){14-16}\cmidrule(lr){17-19}
    Domain ($T=0$) & MTP & \ttt & \hsm & ESR & MTP & \ttt & \hsm & ESR
    & MTP & \ttt & \hsm & ESR & \ttt & \hsm & ESR & \ttt & \hsm & ESR \\
    \midrule
    Coding & 3.39 & 4.36 & 4.18 & 4.35 & 3.49 & 3.93 & 3.84 & 3.77 & 2.39 & 3.08 & 3.58 & 3.57 & 2.75 & 2.65 & 2.64 & 4.11 & 3.91 & 3.88 \\
    Humanities & 3.17 & 3.64 & 3.93 & 3.67 & 3.17 & 3.55 & 3.48 & 3.42 & 2.27 & 2.81 & 3.15 & 3.13 & 2.06 & 2.09 & 2.03 & 2.62 & 2.56 & 2.52 \\
    Math & 3.56 & 4.50 & 4.43 & 4.59 & 3.53 & 4.07 & 3.94 & 3.92 & 2.51 & 3.32 & 3.60 & 3.65 & 2.39 & 2.32 & 2.31 & 3.76 & 3.65 & 3.62 \\
    Multilingual & 2.71 & 3.47 & 3.24 & 3.37 & 3.28 & 3.75 & 3.67 & 3.59 & 2.17 & 2.68 & 3.06 & 3.11 & 2.36 & 2.28 & 2.27 & 3.07 & 2.97 & 2.90 \\
    QA & 2.62 & 3.18 & 3.02 & 3.13 & 3.01 & 3.39 & 3.31 & 3.31 & 2.16 & 2.77 & 3.03 & 3.12 & 2.22 & 2.20 & 2.12 & 2.52 & 2.51 & 2.44 \\
    RAG & 3.13 & 3.85 & 3.69 & 3.86 & 3.04 & 3.42 & 3.38 & 3.30 & 2.39 & 2.97 & 3.49 & 3.50 & 2.39 & 2.28 & 2.32 & 3.19 & 3.12 & 3.08 \\
    Reasoning & 3.13 & 3.78 & 3.66 & 3.67 & 3.30 & 3.69 & 3.59 & 3.56 & 2.34 & 2.92 & 3.25 & 3.30 & 2.36 & 2.36 & 2.33 & 3.25 & 3.19 & 3.15 \\
    Roleplay & 2.14 & 2.36 & 2.28 & 2.34 & 2.35 & 2.52 & 2.50 & 2.45 & 2.25 & 2.64 & 3.08 & 3.19 & 1.80 & 1.81 & 1.80 & 1.82 & 1.86 & 1.85 \\
    STEM & 3.31 & 3.98 & 4.18 & 4.16 & 3.29 & 3.73 & 3.65 & 3.67 & 2.27 & 2.87 & 3.19 & 3.23 & 2.22 & 2.17 & 2.16 & 3.07 & 3.03 & 3.01 \\
    Summarization & 2.55 & 3.03 & 2.87 & 2.88 & 2.65 & 2.89 & 2.86 & 2.85 & 2.22 & 2.71 & 3.20 & 3.17 & 2.21 & 2.18 & 2.18 & 2.62 & 2.57 & 2.55 \\
    Writing & 2.78 & 3.25 & 3.14 & 3.23 & 2.64 & 2.83 & 2.82 & 2.80 & 1.94 & 2.22 & 2.61 & 2.57 & 2.12 & 2.11 & 2.13 & 2.35 & 2.36 & 2.33 \\
    \midrule
    \textbf{AL} ($T=0$) & 2.95 & 3.58 & 3.51 & 3.57 & 3.07 & 3.43 & 3.37 & 3.33 & 2.27 & 2.82 & 3.20 & 3.23 & 2.26 & 2.22 & 2.21 & 2.94 & 2.88 & 2.85 \\
    \midrule
    $\mathbf{AL}_{\mathrm{greedy}}$ ($T=1$) & 2.14 & 2.39 & 2.33 & 2.41 & 2.70 & 2.93 & 2.88 & 2.86 & 2.07 & 2.48 & 2.78 & 2.79 & 2.12 & 2.09 & 2.08 & 2.80 & 2.75 & 2.73 \\
    \midrule
    $\mathbf{AL}_{\mathrm{prob.}}$ ($T=1$) & 2.12 & 2.75 & 2.41 & 2.71 & 2.67 & 3.01 & 2.92 & 2.89 & 1.73 & 2.02 & 2.40 & 2.43 & 1.96 & 1.98 & 1.97 & 2.46 & 2.44 & 2.41 \\
    \bottomrule
  \end{tabular}%
  }
  \caption{Mean EAGLE-3 AL on the SPEED-Bench qualitative split at draft length seven
  for five target models. \hsm denotes uniform layer sampling;
  ESR denotes exponential sparse replacement
  (Section~\ref{sec:hsm-mixing-strategies}). MTP denotes a standard
  training baseline without TTT/HSM/ESR and is available for the first three models.
  \emph{greedy} and \emph{prob.} denote greedy and
  probabilistic drafting, respectively.}
  \label{tab:eagle_vs_hsm_speed}
\end{table}

\textbf{Experimental setup.}
We compare \ttt, uniform \hsm, and \hsm with exponential sparse replacement
(ESR) for training EAGLE-3 drafters. 
We use four MTP/TTT steps per training forward pass for each method. The training runs span eight epochs
on the V2 data. Within each method, checkpoints
are selected by the highest mean AL at $T=0$ on the evaluation benchmark itself.
The comparison covers Llama-3.1-8B-Instruct~\citep{grattafiori2024llama}, Qwen3-8B and Qwen3-235B-A22B-Instruct-2507~\citep{qwen3technicalreport},
Qwen3.5-122B-A10B~\citep{qwen3.5}, and GPT-OSS-120B~\citep{openai2025gptoss120bgptoss20bmodel}.
Standard MTP baselines, trained without HSM or TTT, are available
for three of the five models.
Evaluation follows Section~\ref{sec:evaluation-setup}.

\begin{figure}[t]
  \centering
  \includegraphics[width=\linewidth]{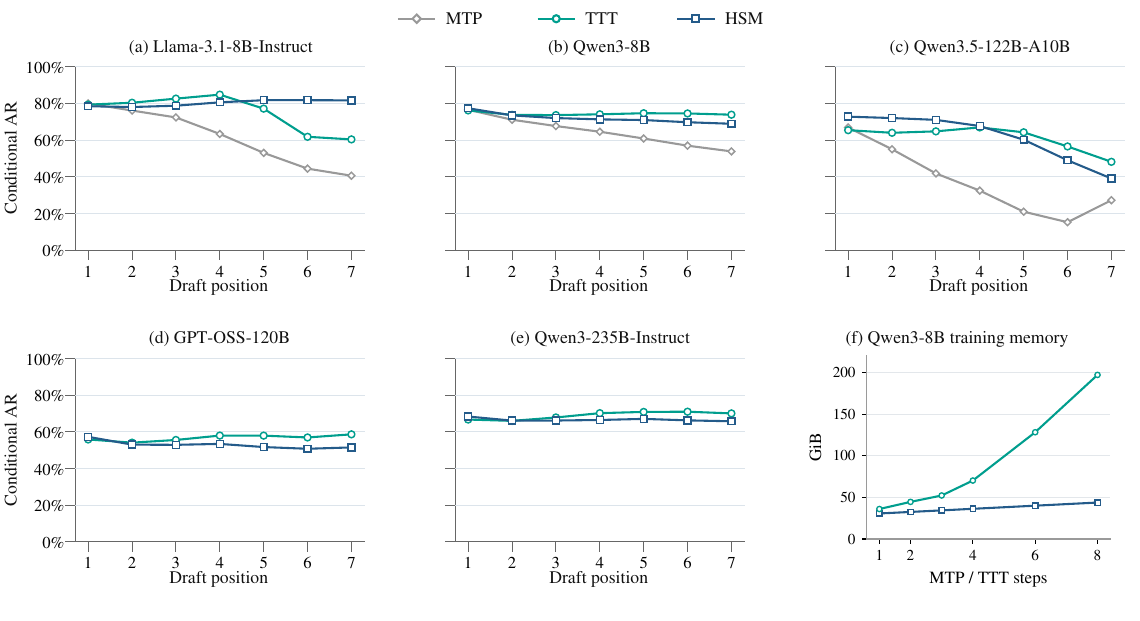}
  \caption{EAGLE-3 training comparisons. (a--e) Acceptance rates at $T=0$
  with greedy decoding for all five target models in
  Table~\ref{tab:eagle_vs_hsm_speed}. MTP baselines are available only
  for the first three models. (f) Peak reserved memory per GPU during
  training EAGLE-3 for Qwen3-8B.}
  \label{fig:eagle-main-ar-t0}
  \label{fig:ttt-memory}
\end{figure}

\textbf{Results.}
TTT and both HSM variants improve ALs over standard MTP
where baselines are available
(Table~\ref{tab:eagle_vs_hsm_speed}). HSM and ESR achieve mean ALs
within 2\% of those achieved by TTT in most settings, while outperforming TTT on Qwen3.5 122B A10B across all reported settings.
Figure~\ref{fig:eagle-main-ar-t0}~(a--e) shows the $T=0$ acceptance rates
for all five models. Acceptance rates at $T=1$ with greedy and
probabilistic drafting are reported in
Appendix~\ref{app:eagle-ar-t1}.
Figure~\ref{fig:eagle-main-ar-t0}(a) compares the
acceptance rates for Llama-3.1-8B-Instruct. Interestingly, \ttt attains higher acceptance rates at the first four
positions, but its acceptance rates decline after the fourth position. In contrast,
\hsm maintains a nearly flat acceptance rate curve and higher acceptance rates
at positions five through seven.

\textbf{Training memory.}
Figure~\ref{fig:ttt-memory}(f) compares peak reserved memory per GPU in
ModelOpt on Qwen3-8B using a single four-GPU GB300 node, aggregated over
1,000 optimizer updates. With four MTP/TTT
steps per training forward pass, \ttt uses 70.3\,GiB compared with 36.3\,GiB for \hsm. Increasing
the number of MTP/TTT steps to eight raises these values to 197.2\,GiB and
43.9\,GiB, respectively. 
Overall, \hsm achieves competitive acceptance lengths
to \ttt while requiring substantially less training memory. 

\FloatBarrier
\newpage
\section{Discussion and Conclusions}
\label{sec:discussion-conclusion}

We studied how to train autoregressive tail-augmented drafters, with a focus on \mtp, to improve acceptance rates across draft positions. Parameter sharing addresses the mismatch between position-specific training and recurrent inference, while hidden state exposure remains a separate challenge. \hsm exposes the drafter to mixtures of aligned hidden states during training, and \hsgd regularizes gradient flow through the recurrent hidden state and embedding pathways. Both methods operate only during training and add no inference cost.

Our co-training experiments favor shared \mtp parameters and MTP supervision from the beginning of pretraining. Following pretraining and supervised fine-tuning, \hsgd yields the highest acceptance rates at early draft positions, while \hsm provides higher tail acceptance rates and the highest mean AL. Combining the methods does not improve mean AL over \hsm alone.
In frozen-backbone MTP boosting, \hsm improves acceptance beyond distillation alone, with the largest gains in acceptance rates at later draft positions. The complete boosting recipes improve mean AL across four Nemotron models and DeepSeek-V4-Pro. For EAGLE-3 drafters, \hsm achieves acceptance lengths close to \ttt in most settings and outperforms it on one of the tested models. The Qwen3-8B memory comparison further shows substantially lower training-memory growth with \hsm than with \ttt in our ModelOpt setup. This motivates its use when co-training \mtp with the target LLM.

The appropriate training method therefore depends on whether the backbone is co-trained or frozen and on the intended serving regime.
High-throughput or larger-batch serving may be dominated by short drafts, so early-position acceptance rates matter more. Low-latency or batch-size-one serving may benefit from longer drafts, so tail acceptance rates matter more. This motivates training separate \mtp drafters for different regions of the latency--throughput Pareto frontier: for example, a drafter co-trained with \hsgd for shorter drafts and one trained with \hsm for longer drafts. In the future, inference engines could support these operating points by dynamically switching between the corresponding \mtp weights according to the selected draft length and workload.

\newpage

\bibliography{references}
\bibliographystyle{references}

\appendix

\Needspace{0.8\textheight}

\section{Additional Results and Experimental Details}
\label{app:additional-results}

\Needspace{0.62\textheight}

\subsection{Pretraining Model Architecture}
\label{app:pretraining-architecture}

Table~\ref{tab:nano-pretraining-architecture} outlines the  Nano-style backbone used for the parameter-sharing and supervision-timing studies, and the pretraining-plus-SFT comparison in Section~\ref{sec:pretraining} and Section~\ref{sec:pretraining-sft-experiments}, respectively.

  \begin{table}[H]
      \centering
  \setlength{\tabcolsep}{3pt}
      \begin{tabular}{@{}ll@{}}
          \toprule
          \textbf{Component} & \textbf{Configuration} \\
          \midrule
          Backbone depth & 54 blocks \\
          Model dimension & 2,688 \\
          Backbone composition & 23 Mamba-2, 8 attention, 23 MoE \\
          Attention heads & 32 query, 8 key--value \\
          Attention head dimension & 128 \\
          Mamba heads & 64 \\
          Routed experts & 128 \\
          Experts selected per token & 6 \\
          Routed-expert dimension & 1,856 \\
          Shared-expert dimension & 3,712 \\
          Router & Learned sigmoid router \\
          Activation & Squared ReLU \\
          Normalization & RMSNorm \\
          Positional embeddings & None \\
          Attention/hidden dropout & 0 \\
          Linear biases & None \\
          Input/output embeddings & Untied \\
          \bottomrule
      \end{tabular}
      \caption{
          Backbone architecture used for the training-from-scratch experiments. The model
          follows the Nemotron 3 Nano hybrid Mamba--Transformer MoE design
          but uses the experimental configuration shown here.
      }
      \label{tab:nano-pretraining-architecture}
  \end{table}

\subsection{Downstream Performance after Pretraining}
\label{app:hsm_downstream}

\paragraph{Evaluation protocol} We evaluate knowledge and factual reasoning using MMLU, including its STEM subset~\citep{hendrycks2021mmlu}, and MMLU-Pro~\citep{wang2024mmlupro}. Code generation is evaluated using MBPP Sanitized~\citep{austin2021program} and HumanEval~\cite{chen2021code}, while mathematical reasoning is measured using MATH-500~\citep{hendrycks2021math,lightman2023verify} and GSM8K~\citep{cobbe2021verifiers}. We evaluate reading comprehension and commonsense reasoning using RACE~\citep{lai2017race}, ARC-Challenge~\citep{clark2018arc}, and WinoGrande~\citep{sakaguchi2020winogrande}. Finally, multilingual performance is evaluated using MGSM~\citep{shi2022multilingual} across six languages and Global-MMLU-Lite~\citep{singh2024globalmmlu} across eight languages.

\paragraph{Results} We compare the downstream performance of Nemotron 3 Nano-style
models after 1T tokens of pretraining, before long-context continuation
and supervised fine-tuning. All models use
shared-parameter MTP with two training steps and follow the controlled
co-training setup described in the main text, differing only in whether
HSM or HSGD is enabled.
Table~\ref{tab:hsm_downstream} reports the downstream evaluation
results across 17 benchmarks. The mean benchmark score is 
$69.42\%$ for the shared MTP baseline, $69.27\%$ for HSM, and $69.84\%$ for HSGD. 
These results demonstrate that enabling HSM or HSGD maintains comparable downstream performance without compromising model quality.

\begin{table}[t]
    \centering
    \small
    \begin{tabular}{lccc}
        \toprule
        Benchmark / metric & Shared MTP & HSM & HSGD \\
        \midrule
        MMLU-Pro Mini (CoT, 5-shot) & 43.56 & 45.64 & 42.23 \\
        MMLU                       & 67.40 & 66.75 & 67.12\\
        MMLU STEM                  & 60.29 & 58.83 & 59.25\\
        HumanEval (pass@1)         & 50.37 & 52.26 & 49.12\\
        MBPP Sanitized {\tiny (pass@1, 3-shot)} &61.75&61.71& 62.68\\
        MATH-500                   & 64.75   & 63.20 & 63.35\\
        GSM8K (CoT)                & 82.34 & 82.94 & 84.61\\
        RACE                       & 81.53 & 80.57 & 80.48\\
        ARC-Challenge (normalized) & 82.08 & 82.51 & 82.00\\
        WinoGrande                 & 72.93 & 73.64 & 73.4\\
        MGSM (German)              & 80.80 & 78.00 & 81.6\\
        MGSM (Spanish)             & 84.80 & 82.80 & 84.4\\
        MGSM (French)              & 73.20 & 74.40 & 75.2\\
        MGSM (Japanese)            & 58.80 & 63.60 & 60.8\\
        MGSM (Russian)             & 79.20 & 76.40 & 81.6\\
        MGSM (Chinese)             & 74.40 & 73.20 & 76.8\\
        Global MMLU-Lite         & 61.97 & 61.22 & 62.69\\
        \midrule
        Mean  & 69.42  & 69.27 & \textbf{69.84}\\
        \bottomrule
    \end{tabular}
    \caption{Downstream performance after 1T tokens of pretraining for Nemotron 3 Nano-style models co-trained with shared-parameter MTP, comparing the baseline against HSM and HSGD. Scores are percentages. Higher is better. HumanEval uses the EvalPlus sanitized evaluation.}
    \label{tab:hsm_downstream}
\end{table}

\FloatBarrier

\subsection{HSM and HSGD through Pretraining and Supervised Fine-Tuning (SFT)}
\label{app:pretraining-sft-results}
Tables~\ref{tab:pretraining-sft-domain-al}
and~\ref{tab:conditional-acceptance-by-draft-length}
report per-domain ALs and per-position acceptance rates for the
final SFT checkpoints compared in
Section~\ref{sec:pretraining-sft-experiments}, evaluated at
$T=0$ and draft length 11.

\begin{table}[H]
  \centering
  \small
  \setlength{\tabcolsep}{4pt}
  \renewcommand{\arraystretch}{1.0}
  \begin{tabular}{@{}lcccc@{}}
    \toprule
    Domain & Shared MTP & \hsm & HSGD & \hsm+HSGD \\
    \midrule
    Coding        & 5.50 & 5.63 & 5.30 & \textbf{5.69} \\
    Humanities    & 5.61 & \textbf{6.77} & 5.75 & 6.56\\
    Math          & 5.77 & \textbf{6.31} & 5.46 & 6.15\\
    Multilingual  & 5.73 & \textbf{5.83} & 5.28 & 5.58\\
    QA            & 4.54 & \textbf{4.83} & 4.62 & 4.64\\
    RAG           & 5.09 & \textbf{5.51} & 5.38 & 5.35\\
    Reasoning     & 5.48 & 5.69 & \textbf{5.70} & 5.59\\
    Roleplay      & 4.53 & 4.92 & \textbf{5.24} & 4.95\\
    STEM          & 5.55 & 6.07 & 5.60 & \textbf{6.16}\\
    Summarization & 4.82 & \textbf{4.92} & 4.79 & 4.71\\
    Writing       & 4.44 & 4.74 & 4.51 & \textbf{4.80}\\
    \midrule
    \textbf{Mean AL} & 5.19 & \textbf{5.57} & 5.24 & 5.47\\
    \bottomrule
  \end{tabular}
  \caption{ALs on the SPEED-Bench qualitative split. Bold indicates the highest AL in each row.}
  \label{tab:pretraining-sft-domain-al}
  \end{table}

\begin{table}[H]
  \centering
  \small
  \setlength{\tabcolsep}{2.5pt}
  \renewcommand{\arraystretch}{1.12}
  \begin{tabular}{@{}l*{11}{c}@{}}
    \toprule
    Method & AR@1 & AR@2 & AR@3 & AR@4 & AR@5 & AR@6 & AR@7 & AR@8 & AR@9 & AR@10 & AR@11 \\
    \midrule
    MTP & 86.29 & 85.27 & 84.69 & 84.20 & 83.06 & 81.31 & 79.46 & 73.91 & 73.74 & 70.71 & 68.24 \\
    \hsm & 86.07 & 84.48 & 84.33 & 84.78 & \textbf{85.01} & \textbf{85.65} & \textbf{85.85} & \textbf{85.68} & \textbf{85.49} & \textbf{84.17} & \textbf{83.20} \\
    HSGD & \textbf{87.09} & \textbf{86.12} & \textbf{85.95} & \textbf{85.55} & 83.65 & 79.67 & 77.41 & 73.91 & 72.64 & 67.63 & 66.14 \\
    \hsm+HSGD & 86.05 & 84.17 & 83.91 & 84.42 & 84.40 & 85.05 & 84.50 & 83.77 & 82.74 & 79.79 & 78.74 \\
    \bottomrule
  \end{tabular}
  \caption{Acceptance rate (\%) by draft position on the
  qualitative benchmark. Bold indicates the highest rate at each draft position.}
  \label{tab:conditional-acceptance-by-draft-length}
\end{table}

\FloatBarrier

\subsection{MTP Boosting Setup}
\label{app:boosting-data}
\label{app:boosting-training}

V2 denotes \textit{Nemotron-Post-Training-Dataset-v2}
\citep{NemotronPostTrainingDatasetV2}, and V3 denotes
\textit{Nemotron-RL-Super-Training-Blends}. V2+V3 denotes a mixture of
these two prompt sources. Table~\ref{tab:boosting-recipes} summarizes the
recipes for the boosted checkpoints in Section~\ref{sec:frozen-backbone-mtp-training}.

\begin{table}[!htb]
\centering
\small
\setlength{\tabcolsep}{5pt}
\begin{tabular}{@{}lllcl@{}}
\toprule
Model & Data & Loss & MTP steps & Training iterations \\
\midrule
\super{} MTPv2 & V2 & KL, temperature 2 & 7 & 30k \\
\super3dot5{} & V2+V3 & LK & 7 & 25k (after 4k) \\
\lightning{} & V2+V3 & LK & 7 & 48k \\
\ultra{} & V2+V3 & KL, temperature 2 & 7 & 12k \\
DeepSeek-V4-Pro & V2+V3 & LK & 5 & 39k \\
\bottomrule
\end{tabular}
\caption{Training recipes for the reported boosted checkpoints.
 MTP steps denotes the number of recurrent prediction steps within each
 training iteration. KL temperatures are loss temperatures;
 LK follows~\citet{samarin2026lk}. \super3dot5{} receives 25k iterations on
 the final backbone after an earlier 4k boosting stage on an SFT checkpoint.}
\label{tab:boosting-recipes}
\end{table}

\textbf{Training configurations.}
The four Nemotron MTP boosting configurations in Table~\ref{tab:boosting-recipes}
use seven recurrent MTP steps with shared parameters, AdamW, a global
batch size of 64, and a sequence length of 8,192 tokens. They use
microbatches of one sequence and BF16 precision.
For \super{}, we initialize from the official released MTP checkpoint.
For \super3dot5{}, we initialize from an MTP module previously boosted
on an SFT checkpoint and continue boosting on the final backbone.
Table~\ref{tab:boosting-optimizer} summarizes the optimizer and learning-rate
settings. All five models use Adam $\epsilon=10^{-8}$ and gradient
clipping with a maximum norm of 1.0; warmup is linear from zero.

\begin{table}[!htb]
\centering
\small
\setlength{\tabcolsep}{4pt}
\begin{tabular}{@{}l c l l c c@{}}
\toprule
Model & Peak LR & Schedule & Warmup & $(\beta_1,\beta_2)$ & Weight decay \\
\midrule
\super{} & $10^{-5}$ & Constant & 30k samples & $(0.9,0.95)$ & 0.1 \\
\super3dot5{} & $10^{-5}$ & Constant & 30k samples & $(0.9,0.95)$ & 0.1 \\
\lightning{} & $10^{-5}$ & Constant & 30k samples & $(0.9,0.98)$ & 0 \\
\ultra{} & $10^{-5}$ & Cosine & 7.5k samples & $(0.9,0.95)$ & 0.1 \\
DeepSeek-V4-Pro & $10^{-4}$ & Linear to zero & 5 updates & $(0.9,0.999)$ & 0 \\
\bottomrule
\end{tabular}
\caption{Optimizer hyperparameters for MTP boosting.}
\label{tab:boosting-optimizer}
\end{table}

\textbf{Training data.} We used V2 and V3 data. With V2, 1.4 million seed prompts are drawn from the
non-multilingual categories: chat, code, math, and STE. For V3 we exclude the \texttt{swe2} split and separated each user turn together with its preceding conversation to a separate prompt as long as the included successive user turns do not exceed 7K tokens. For each boosted model we generated on-policy rollouts with the corresponding backbone at generation
temperature $\tau=1$. Training sequences are capped at 8K tokens, and the loss
is accumulated only over assistant-response tokens.

For the released \super{} MTPv2 we used seed from only V2. Puzzle adaptation reuses these \super{}-generated V2 responses;
initialization and adaptation details appear in Appendix~\ref{app:puzzle-adaptation}. For any other boosted model we used a mixture of V2+V3.

\subsection{MTP Boosting Results by Domain}
\label{app:boosting-domains}

Table~\ref{tab:mtp-boosting-domains} reports AL across the 11 domains of
the SPEED-Bench qualitative split for \super{}, \super3dot5{},
\lightning{}, \ultra{}, and DeepSeek V4 Pro. Columns are grouped by model and evaluation
temperature. Pre and Post denote the pre-boosting and boosted checkpoints
reported in Table~\ref{tab:mtp-boosting-summary}, respectively.

\begin{table}[!htp]
\centering
\small
\setlength{\tabcolsep}{1.5pt}
\resizebox{\linewidth}{!}{%
\begin{tabular}{@{}l*{20}{r}@{}}
\toprule
 & \multicolumn{4}{c}{\shortstack{Nemotron 3\\Super}} & \multicolumn{4}{c}{\shortstack{Nemotron 3.5\\Super}} & \multicolumn{4}{c}{\shortstack{Nemotron 3.5\\Lightning}} & \multicolumn{4}{c}{\shortstack{Nemotron 3\\Ultra}} & \multicolumn{4}{c}{\shortstack{DeepSeek-\\V4-Pro}} \\
\cmidrule(lr){2-5}\cmidrule(lr){6-9}\cmidrule(lr){10-13}\cmidrule(lr){14-17}\cmidrule(lr){18-21}
 & \multicolumn{2}{c}{$T=0$} & \multicolumn{2}{c}{$T=1$} & \multicolumn{2}{c}{$T=0$} & \multicolumn{2}{c}{$T=1$} & \multicolumn{2}{c}{$T=0$} & \multicolumn{2}{c}{$T=1$} & \multicolumn{2}{c}{$T=0$} & \multicolumn{2}{c}{$T=1$} & \multicolumn{2}{c}{$T=0$} & \multicolumn{2}{c}{$T=1$} \\
\cmidrule(lr){2-3}\cmidrule(lr){4-5}\cmidrule(lr){6-7}\cmidrule(lr){8-9}\cmidrule(lr){10-11}\cmidrule(lr){12-13}\cmidrule(lr){14-15}\cmidrule(lr){16-17}\cmidrule(lr){18-19}\cmidrule(lr){20-21}
Domain & Pre & Post & Pre & Post & Pre & Post & Pre & Post & Pre & Post & Pre & Post & Pre & Post & Pre & Post & Pre & Post & Pre & Post \\
\midrule
Coding & 3.78 & 4.77 & 3.47 & \multicolumn{1}{r|}{4.73} & 4.97 & 5.56 & 4.34 & \multicolumn{1}{r|}{5.44} & 3.87 & 4.36 & 3.55 & \multicolumn{1}{r|}{4.23} & 5.15 & 5.45 & 4.86 & \multicolumn{1}{r|}{5.35} & 2.86 & 4.48 & 2.74 & 4.40 \\
Humanities & 3.26 & 4.16 & 2.84 & \multicolumn{1}{r|}{3.83} & 3.70 & 3.98 & 3.25 & \multicolumn{1}{r|}{3.82} & 3.09 & 3.41 & 2.78 & \multicolumn{1}{r|}{3.24} & 3.95 & 4.10 & 3.63 & \multicolumn{1}{r|}{4.06} & 2.80 & 4.33 & 2.62 & 3.75 \\
Math & 3.73 & 4.69 & 3.35 & \multicolumn{1}{r|}{4.50} & 4.53 & 4.84 & 4.09 & \multicolumn{1}{r|}{4.83} & 3.98 & 4.32 & 3.57 & \multicolumn{1}{r|}{4.11} & 5.13 & 5.34 & 4.96 & \multicolumn{1}{r|}{5.38} & 2.96 & 4.54 & 2.85 & 4.45 \\
Multilingual & 4.05 & 5.34 & 3.58 & \multicolumn{1}{r|}{5.10} & 4.95 & 5.55 & 4.27 & \multicolumn{1}{r|}{5.36} & 4.12 & 4.63 & 3.66 & \multicolumn{1}{r|}{4.40} & 5.14 & 5.38 & 4.85 & \multicolumn{1}{r|}{5.37} & 2.84 & 4.49 & 2.73 & 4.38 \\
QA & 3.16 & 3.87 & 2.82 & \multicolumn{1}{r|}{3.86} & 4.08 & 4.53 & 3.62 & \multicolumn{1}{r|}{4.38} & 3.13 & 3.47 & 2.82 & \multicolumn{1}{r|}{3.34} & 4.25 & 4.47 & 3.82 & \multicolumn{1}{r|}{4.29} & 2.79 & 4.31 & 2.70 & 4.06 \\
RAG & 3.78 & 4.79 & 3.39 & \multicolumn{1}{r|}{4.71} & 5.05 & 5.54 & 4.42 & \multicolumn{1}{r|}{5.50} & 3.85 & 4.27 & 3.58 & \multicolumn{1}{r|}{4.22} & 5.21 & 5.38 & 4.93 & \multicolumn{1}{r|}{5.34} & 2.98 & 4.86 & 2.84 & 4.67 \\
Reasoning & 3.59 & 4.49 & 3.26 & \multicolumn{1}{r|}{4.37} & 4.31 & 4.79 & 3.92 & \multicolumn{1}{r|}{4.75} & 3.57 & 3.96 & 3.23 & \multicolumn{1}{r|}{3.84} & 4.67 & 4.90 & 4.45 & \multicolumn{1}{r|}{4.90} & 2.84 & 4.39 & 2.73 & 4.28 \\
Roleplay & 2.82 & 3.33 & 2.46 & \multicolumn{1}{r|}{3.21} & 3.44 & 3.78 & 3.05 & \multicolumn{1}{r|}{3.70} & 2.87 & 3.18 & 2.54 & \multicolumn{1}{r|}{3.00} & 2.80 & 2.94 & 2.57 & \multicolumn{1}{r|}{2.90} & 2.38 & 3.16 & 2.32 & 3.15 \\
STEM & 3.30 & 4.10 & 2.96 & \multicolumn{1}{r|}{4.04} & 3.76 & 4.10 & 3.41 & \multicolumn{1}{r|}{4.05} & 3.30 & 3.59 & 2.98 & \multicolumn{1}{r|}{3.39} & 4.26 & 4.44 & 3.97 & \multicolumn{1}{r|}{4.42} & 2.76 & 4.00 & 2.64 & 3.86 \\
Summarization & 3.48 & 4.31 & 3.24 & \multicolumn{1}{r|}{4.25} & 4.72 & 5.24 & 4.11 & \multicolumn{1}{r|}{5.04} & 3.67 & 4.10 & 3.27 & \multicolumn{1}{r|}{4.01} & 4.41 & 4.55 & 4.08 & \multicolumn{1}{r|}{4.54} & 2.73 & 4.17 & 2.66 & 4.13 \\
Writing & 2.99 & 3.66 & 2.71 & \multicolumn{1}{r|}{3.56} & 3.65 & 4.00 & 3.18 & \multicolumn{1}{r|}{3.87} & 2.77 & 3.03 & 2.51 & \multicolumn{1}{r|}{2.93} & 3.29 & 3.48 & 3.01 & \multicolumn{1}{r|}{3.44} & 2.53 & 3.44 & 2.43 & 3.39 \\
\midrule
\textbf{Mean} & 3.45 & \textbf{4.32} & 3.10 & \multicolumn{1}{r|}{\textbf{4.20}} & 4.29 & \textbf{4.72} & 3.79 & \multicolumn{1}{r|}{\textbf{4.61}} & 3.48 & \textbf{3.85} & 3.13 & \multicolumn{1}{r|}{\textbf{3.70}} & 4.39 & \textbf{4.58} & 4.10 & \multicolumn{1}{r|}{\textbf{4.55}} & 2.77 & \textbf{4.20} & 2.66 & \textbf{4.05} \\
\bottomrule
\end{tabular}}
\caption{Acceptance length by domain before and after MTP boosting on the SPEED-Bench qualitative split with draft length seven. Pre and Post denote the pre-boosting and boosted checkpoints, respectively. Evaluation at $T=1$ uses probabilistic draft sampling.}
\label{tab:mtp-boosting-domains}
\end{table}

\newpage
\subsection{MTP Boosting Acceptance Rates by Draft Position}
\label{app:boosting-ar}

Table~\ref{tab:mtp-boosting-ar} reports the acceptance rates
shown in Figure~\ref{fig:mtp-boosting-ar}(a) for the pre-boosting and boosted
checkpoints.

\begin{table}[!htp]
  \centering
  \small
  \setlength{\tabcolsep}{3pt}
  \begin{tabular}{@{}clrrrrrrr@{}}
    \toprule
    $T$ & Checkpoint & AR@1 & AR@2 & AR@3 & AR@4 & AR@5 & AR@6 & AR@7 \\
    \midrule
    \multicolumn{9}{l}{\textbf{\super{}}} \\
    \midrule
    0 & Pre boosting & 0.81 & 0.74 & 0.70 & 0.59 & 0.54 & 0.44 & 0.41 \\
    0 & Boosted & \textbf{0.83} & \textbf{0.79} & \textbf{0.77} & \textbf{0.76} & \textbf{0.75} & \textbf{0.75} & \textbf{0.75} \\
    1 & Pre boosting & 0.79 & 0.68 & 0.63 & 0.49 & 0.44 & 0.34 & 0.31 \\
    1 & Boosted & \textbf{0.82} & \textbf{0.78} & \textbf{0.76} & \textbf{0.75} & \textbf{0.74} & \textbf{0.73} & \textbf{0.73} \\
    \midrule
    \multicolumn{9}{l}{\textbf{\super3dot5{}}} \\
    \midrule
    0 & Pre boosting & 0.83 & 0.80 & 0.77 & 0.73 & 0.71 & 0.68 & 0.63 \\
    0 & Boosted & \textbf{0.84} & \textbf{0.81} & \textbf{0.79} & \textbf{0.79} & \textbf{0.79} & \textbf{0.79} & \textbf{0.79} \\
    1 & Pre boosting & 0.82 & 0.76 & 0.72 & 0.66 & 0.61 & 0.57 & 0.51 \\
    1 & Boosted & \textbf{0.84} & \textbf{0.80} & \textbf{0.79} & \textbf{0.78} & \textbf{0.77} & \textbf{0.77} & \textbf{0.78} \\
    \midrule
    \multicolumn{9}{l}{\textbf{\lightning{}}} \\
    \midrule
    0 & Pre boosting & 0.74 & 0.72 & 0.70 & 0.69 & 0.68 & 0.69 & 0.69 \\
    0 & Boosted & \textbf{0.76} & \textbf{0.75} & \textbf{0.74} & \textbf{0.74} & \textbf{0.75} & \textbf{0.76} & \textbf{0.77} \\
    1 & Pre boosting & 0.71 & 0.68 & 0.66 & 0.64 & 0.63 & 0.63 & 0.64 \\
    1 & Boosted & \textbf{0.75} & \textbf{0.73} & \textbf{0.72} & \textbf{0.73} & \textbf{0.73} & \textbf{0.74} & \textbf{0.76} \\
    \midrule
    \multicolumn{9}{l}{\textbf{\ultra{}}} \\
    \midrule
    0 & Pre boosting & 0.82 & 0.79 & 0.78 & 0.77 & 0.76 & 0.76 & 0.76 \\
    0 & Boosted & \textbf{0.83} & \textbf{0.80} & \textbf{0.79} & \textbf{0.79} & \textbf{0.79} & \textbf{0.80} & \textbf{0.80} \\
    1 & Pre boosting & 0.80 & 0.77 & 0.75 & 0.74 & 0.73 & 0.73 & 0.72 \\
    1 & Boosted & \textbf{0.83} & \textbf{0.80} & \textbf{0.79} & \textbf{0.79} & \textbf{0.79} & \textbf{0.79} & \textbf{0.79} \\
    \midrule
    \multicolumn{9}{l}{\textbf{DeepSeek-V4-Pro}} \\
    \midrule
    0 & Pre boosting & \textbf{0.84} & 0.63 & 0.45 & 0.35 & 0.30 & 0.27 & 0.25 \\
    0 & Boosted & 0.83 & \textbf{0.79} & \textbf{0.76} & \textbf{0.75} & \textbf{0.74} & \textbf{0.74} & \textbf{0.74} \\
    1 & Pre boosting & \textbf{0.84} & 0.60 & 0.41 & 0.29 & 0.24 & 0.22 & 0.21 \\
    1 & Boosted & 0.82 & \textbf{0.77} & \textbf{0.75} & \textbf{0.73} & \textbf{0.72} & \textbf{0.71} & \textbf{0.71} \\
    \bottomrule
  \end{tabular}
  \caption{Acceptance rates before and after MTP boosting on
  the SPEED-Bench qualitative split with draft length seven. Temperatures
  refer to evaluation. Bold marks the highest AR, including
  ties, at each draft position within each model and temperature pair where
  both measurements are available.}
  \label{tab:mtp-boosting-ar}
\end{table}

\newpage
\FloatBarrier

\subsection{MTP Boosting Acceptance Length by Draft Length}
\label{app:boosting-al-by-draft-length}

Table~\ref{tab:boosting-al-by-draft-length} reports the pre-boosting and
post-boosting AL measurements underlying Figure~\ref{fig:mtp-boosting-ar}(b).
Draft length denotes the maximum number of draft tokens per verification
step, rather than the position of an individual token. 

\begin{table}[!htb]
\centering
\small
\setlength{\tabcolsep}{3pt}
\renewcommand{\arraystretch}{1.05}
\begin{tabular}{@{}clrrrrrrr@{}}
\toprule
$T$ & Checkpoint & DL1 & DL2 & DL3 & DL4 & DL5 & DL6 & DL7 \\
\midrule
\multicolumn{9}{l}{\textbf{\super{}}} \\
\midrule
0 & Pre & 1.86 & 2.52 & 2.99 & 3.26 & 3.39 & 3.46 & 3.45 \\
0 & Post & 1.88 & 2.60 & 3.17 & 3.60 & 3.91 & 4.15 & 4.32 \\
1 & Pre & 1.84 & 2.43 & 2.81 & 2.99 & 3.07 & 3.09 & 3.10 \\
1 & Post & 1.87 & 2.58 & 3.13 & 3.54 & 3.81 & 4.01 & 4.20 \\
\midrule
\multicolumn{9}{l}{\textbf{\super3dot5{}}} \\
\midrule
0 & Pre & 1.89 & 2.64 & 3.22 & 3.66 & 3.95 & 4.16 & 4.29 \\
0 & Post & 1.90 & 2.68 & 3.31 & 3.80 & 4.19 & 4.49 & 4.72 \\
1 & Pre & 1.87 & 2.57 & 3.08 & 3.41 & 3.61 & 3.71 & 3.79 \\
1 & Post & 1.90 & 2.65 & 3.27 & 3.74 & 4.11 & 4.38 & 4.61 \\
\midrule
\multicolumn{9}{l}{\textbf{DeepSeek-V4-Pro}} \\
\midrule
0 & Pre & 1.87 & 2.42 & 2.66 & 2.74 & 2.76 & 2.76 & 2.77 \\
0 & Post & 1.87 & 2.57 & 3.12 & 3.51 & 3.82 & 4.03 & 4.20 \\
1 & Pre & 1.86 & 2.38 & 2.59 & 2.65 & 2.66 & 2.67 & 2.66 \\
1 & Post & 1.86 & 2.55 & 3.07 & 3.45 & 3.74 & 3.93 & 4.05 \\
\midrule
\multicolumn{9}{l}{\textbf{\lightning{}}} \\
\midrule
0 & Pre & 1.81 & 2.42 & 2.85 & 3.13 & 3.31 & 3.43 & 3.48 \\
0 & Post & 1.83 & 2.48 & 2.96 & 3.31 & 3.56 & 3.74 & 3.85 \\
1 & Pre & 1.78 & 2.33 & 2.68 & 2.90 & 3.02 & 3.10 & 3.13 \\
1 & Post & 1.82 & 2.44 & 2.89 & 3.21 & 3.43 & 3.58 & 3.70 \\
\midrule
\multicolumn{9}{l}{\textbf{\ultra{}}} \\
\midrule
0 & Pre & 1.88 & 2.61 & 3.20 & 3.66 & 4.00 & 4.27 & 4.39 \\
0 & Post & 1.89 & 2.63 & 3.23 & 3.71 & 4.08 & 4.35 & 4.58 \\
1 & Pre & 1.86 & 2.56 & 3.09 & 3.48 & 3.75 & 3.95 & 4.10 \\
1 & Post & 1.88 & 2.62 & 3.21 & 3.69 & 4.04 & 4.34 & 4.55 \\
\bottomrule
\end{tabular}
\caption{Acceptance length before and after MTP boosting on the
SPEED-Bench qualitative split at draft lengths one through
seven. $T=0$ uses greedy drafting and $T=1$ uses probabilistic drafting.}
\label{tab:boosting-al-by-draft-length}
\end{table}

\newpage
\FloatBarrier

\subsection{MTP Boosting on a Distilled Backbone}
\label{app:puzzle-adaptation}
We test whether \super{}'s MTP module can be reused on
Nemotron-Labs-3-Puzzle-75B-A9B, a compressed derivative of \super{}
trained to preserve \super{}'s behavior through distillation
\citep{bercovich2026nemotronlabs3puzzle}.

The original \super{} MTP module achieves comparable AL on both
backbones without additional training. To improve acceptance, we
initialize from the \hsm-boosted \super{} MTP module and continue
boosting on the frozen Nemotron-Labs-3-Puzzle-75B-A9B backbone using the same
\super{}-generated V2 responses (Appendix~\ref{app:boosting-data}).
The adapted drafter achieves a mean AL of 4.34.

Table~\ref{tab:super-turbo-mtp-domain} reports per-domain ALs for the
adapted drafter, with the original MTP evaluated on \super{} as
a reference.
\begin{table}[H]
  \centering
  \setlength{\tabcolsep}{3pt}
  \small
  \begin{tabular}{@{}lcc@{}}
    \toprule
    Domain & \shortstack{Original MTP on\\ \super{}} & \shortstack{Adapted MTP on\\Nemotron-Labs-3-Puzzle-75B-A9B} \\
    \midrule
    Coding        & 3.78 & 4.80 \\
    Humanities    & 3.26 & 4.14 \\
    Math          & 3.73 & 4.75 \\
    Multilingual  & 4.05 & 5.31 \\
    QA            & 3.16 & 3.95 \\
    RAG           & 3.78 & 4.78 \\
    Reasoning     & 3.59 & 4.46 \\
    Roleplay      & 2.82 & 3.45 \\
    STEM          & 3.30 & 4.15 \\
    Summarization & 3.48 & 4.22 \\
    Writing       & 2.99 & 3.73 \\
    \midrule
    \textbf{Mean} & 3.45 & \textbf{4.34} \\
    \bottomrule
  \end{tabular}
  \caption{Per-domain AL for the original \super{} MTP evaluated on
  \super{} and the adapted MTP evaluated on Nemotron-Labs-3-Puzzle-75B-A9B,
  using the SPEED-Bench qualitative split with draft length seven.}
  \label{tab:super-turbo-mtp-domain}
\end{table}

\FloatBarrier

\subsection{MTP Boosting with HSM and HSGD}
\label{app:hsm-hsgd-boosting-ar}

Table~\ref{tab:hsm-hsgd-boosting-ar} reports the acceptance rates plotted
in Figure~\ref{fig:hsm-hsgd-boosting-ar}, comparing the pre-boosting checkpoint
with HSM, HSGD, and their combination at draft length 11.

\begin{table}[!htb]
    \centering
    \small
    \setlength{\tabcolsep}{4pt}
    \resizebox{\textwidth}{!}{%
    \begin{tabular}{l*{11}{r}}
      \toprule
      Checkpoint & AR@1 & AR@2 & AR@3 & AR@4 & AR@5 & AR@6 &
      AR@7 & AR@8 & AR@9 & AR@10 & AR@11 \\
      \midrule
      Pre boosting
        & 81.37 & 73.65 & 70.02 & 58.42 & 53.90 & 44.22
        & 40.41 & 35.65 & 32.28 & 30.44 & 33.17 \\
      HSM Boosted
        & 81.51 & 77.85 & 75.43 & \textbf{74.09} & \textbf{73.17} & \textbf{72.40}
        & \textbf{71.48} & \textbf{70.81} & \textbf{70.34} & \textbf{69.11} & \textbf{68.91} \\
      HSGD Boosted
        & \textbf{82.37} & 75.96 & 74.54 & 64.89 & 66.60 & 56.13
        & 57.67 & 50.84 & 52.30 & 49.81 & 52.00 \\
      HSGD+HSM Boosted
        & 81.70 & \textbf{78.15} & \textbf{75.65} & 73.93 & 72.82 & 71.56
        & 70.99 & 70.12 & 70.07 & 68.90 & 67.11 \\
      \bottomrule
    \end{tabular}%
    }
    \caption{Acceptance rates (\%) for \super{} before and
after MTP boosting on the SPEED-Bench qualitative split at
$T=0$ and draft length 11.}
    \label{tab:hsm-hsgd-boosting-ar}
  \end{table}

\newpage
\subsection{Effect of HSM on Acceptance}
\label{app:hsm-acceptance-rates}

Table~\ref{tab:super_hsm_ablation} reports the acceptance lengths and
acceptance rates for the KD-only and \hsm+KD checkpoints
shown in Figure~\ref{fig:super-hsm-ablation-ar}. Both configurations use an LK loss and are evaluated after 10k boosting iterations of \super{}'s MTP.

\begin{table}[H]
\centering
\small
\setlength{\tabcolsep}{3pt}
\renewcommand{\arraystretch}{1.12}
\begin{tabular}{@{}l r r r r r r r r@{}}
\hline
Method & AL & AR@1 & AR@2 & AR@3 & AR@4 & AR@5 & AR@6 & AR@7 \\
\hline
\multicolumn{9}{c}{Evaluation $T=0$, greedy drafting} \\
\hline
KD only (LK) & 3.83 & \textbf{0.82} & 0.76 & 0.75 & 0.65 & 0.69 & 0.57 & 0.61 \\
\hsm+KD (LK) & \textbf{4.22} & \textbf{0.82} & \textbf{0.78} & \textbf{0.76} & \textbf{0.75} & \textbf{0.74} & \textbf{0.73} & \textbf{0.72} \\
\hline
\multicolumn{9}{c}{Evaluation $T=1$, probabilistic drafting} \\
\hline
KD only (LK) & 3.58 & \textbf{0.81} & 0.73 & 0.72 & 0.60 & 0.63 & 0.51 & 0.55 \\
\hsm+KD (LK) & \textbf{4.02} & \textbf{0.81} & \textbf{0.77} & \textbf{0.74} & \textbf{0.72} & \textbf{0.71} & \textbf{0.70} & \textbf{0.69} \\
\hline
\end{tabular}
\caption{Effect of \hsm after 10k boosting iterations of \super{}'s
MTP. AR@1--7 denote acceptance rates at the seven draft-token
positions. Bold indicates the highest value.}
\label{tab:super_hsm_ablation}
\end{table}

\FloatBarrier

\subsection{Number of MTP Training Steps at $T=0$ Evaluation}
\label{app:recurrence-depth-t0}

Table~\ref{tab:mtp_recurrence_depth} reports AL and acceptance rates for
\super{}, \lightning{}, and \super3dot5{} at evaluation temperature $T=0$
with greedy drafting and draft length seven. We vary the number of recurrent
\mtp training steps from two to seven.
\super{} uses 4k training iterations on on-policy V2 data generated at
rollout temperature $\tau=1$. \lightning{} and \super3dot5{} follow the default
ablation setup in Appendix~\ref{app:boosting-training}.

\begin{table}[H]
\centering
\small
\setlength{\tabcolsep}{3pt}
\renewcommand{\arraystretch}{1.12}
\begin{tabular}{@{}l r r r r r r r r@{}}
\hline
Configuration & AL & AR@1 & AR@2 & AR@3 & AR@4 & AR@5 & AR@6 & AR@7 \\
\hline
\multicolumn{9}{c}{\super{}} \\
\hline
Pre boosting & 3.45 & 0.81 & 0.74 & 0.70 & 0.59 & 0.54 & 0.44 & 0.41 \\
2 MTP steps & 3.78 & 0.83 & 0.78 & 0.73 & 0.65 & 0.58 & 0.51 & 0.47 \\
3 MTP steps & 4.04 & 0.82 & 0.78 & 0.75 & 0.72 & 0.69 & 0.65 & 0.62 \\
5 MTP steps & 4.12 & 0.82 & 0.78 & 0.75 & 0.73 & 0.72 & 0.70 & 0.69 \\
7 MTP steps & \textbf{4.14} & 0.82 & 0.78 & 0.75 & 0.74 & 0.72 & 0.71 & 0.71 \\
\hline
\multicolumn{9}{c}{\lightning{}} \\
\hline
Pre boosting & 3.48 & 0.74 & 0.72 & 0.70 & 0.69 & 0.68 & 0.69 & 0.69 \\
2 MTP steps & 3.59 & 0.77 & 0.74 & 0.72 & 0.69 & 0.66 & 0.63 & 0.60 \\
3 MTP steps & 3.71 & 0.76 & 0.74 & 0.73 & 0.72 & 0.72 & 0.71 & 0.71 \\
5 MTP steps & 3.75 & 0.76 & 0.74 & 0.73 & 0.73 & 0.73 & 0.73 & 0.75 \\
7 MTP steps & \textbf{3.77} & 0.76 & 0.74 & 0.73 & 0.73 & 0.74 & 0.74 & 0.75 \\
\hline
\multicolumn{9}{c}{\super3dot5{}} \\
\hline
Pre boosting & 4.29 & 0.83 & 0.80 & 0.77 & 0.73 & 0.71 & 0.68 & 0.63 \\
2 MTP steps & 4.30 & 0.84 & 0.81 & 0.77 & 0.73 & 0.68 & 0.63 & 0.58 \\
3 MTP steps & 4.54 & 0.84 & 0.81 & 0.78 & 0.77 & 0.75 & 0.74 & 0.72 \\
5 MTP steps & 4.60 & 0.84 & 0.81 & 0.79 & 0.78 & 0.77 & 0.77 & 0.77 \\
7 MTP steps & \textbf{4.64} & 0.84 & 0.81 & 0.79 & 0.78 & 0.77 & 0.77 & 0.77 \\
\hline
\end{tabular}
\caption{Effect of the number of MTP steps during training with \hsm, evaluated at
temperature $T=0$ with a draft length of seven. \emph{Pre boosting} denotes the
\mtp module before boosting.}
\label{tab:mtp_recurrence_depth}
\end{table}

\FloatBarrier

\subsection{Number of MTP Training Steps at $T=1$ Evaluation}
\label{app:recurrence-depth-t1}

\begin{figure}[!htb]
\centering
\includegraphics[width=0.95\linewidth]{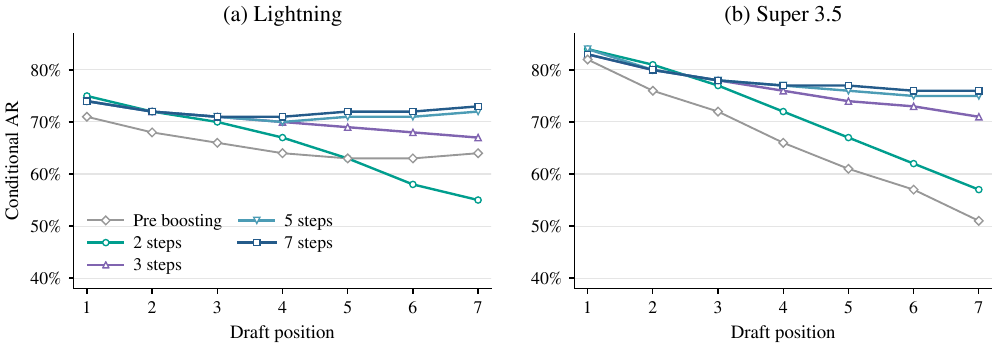}
\caption{Acceptance rates for the MTP-step comparisons in
Table~\ref{tab:mtp_recurrence_depth_t1}: (a) \lightning{} and (b) \super3dot5{}.
Evaluation uses $T=1$, probabilistic drafting, and draft length seven.}
\label{fig:mtp-steps-lightning-t1-conditional-ar}
\end{figure}

Table~\ref{tab:mtp_recurrence_depth_t1} reports AL and acceptance rates for
\lightning{} and \super3dot5{} at evaluation temperature $T=1$
with probabilistic drafting and draft length seven.
Figure~\ref{fig:mtp-steps-lightning-t1-conditional-ar} shows the corresponding acceptance rates.

Additional MTP training steps mainly improve acceptance rates at later draft
positions. The largest AL gain comes from two to three steps, with
diminishing returns thereafter and little change in first-position acceptance rate.

\begin{table}[H]
\centering
\small
\setlength{\tabcolsep}{3pt}
\renewcommand{\arraystretch}{1.12}
\begin{tabular}{@{}l*{8}{r}@{}}
\hline
Configuration & AL & AR@1 & AR@2 & AR@3 & AR@4 & AR@5 & AR@6 & AR@7 \\
\hline
\multicolumn{9}{c}{\lightning{}} \\
\hline
Pre boosting & 3.13 & 0.71 & 0.68 & 0.66 & 0.64 & 0.63 & 0.63 & 0.64 \\
2 MTP steps & 3.39 & 0.75 & 0.72 & 0.70 & 0.67 & 0.63 & 0.58 & 0.55 \\
3 MTP steps & 3.51 & 0.74 & 0.72 & 0.71 & 0.70 & 0.69 & 0.68 & 0.67 \\
5 MTP steps & \textbf{3.56} & 0.74 & 0.72 & 0.71 & 0.70 & 0.71 & 0.71 & 0.72 \\
7 MTP steps & \textbf{3.56} & 0.74 & 0.72 & 0.71 & 0.71 & 0.72 & 0.72 & 0.73 \\
\hline
\multicolumn{9}{c}{\super3dot5{}} \\
\hline
Pre boosting & 3.79 & 0.82 & 0.76 & 0.72 & 0.66 & 0.61 & 0.57 & 0.51 \\
2 MTP steps & 4.22 & 0.84 & 0.81 & 0.77 & 0.72 & 0.67 & 0.62 & 0.57 \\
3 MTP steps & 4.46 & 0.83 & 0.80 & 0.78 & 0.76 & 0.74 & 0.73 & 0.71 \\
5 MTP steps & 4.53 & 0.84 & 0.80 & 0.78 & 0.77 & 0.76 & 0.75 & 0.75 \\
7 MTP steps & \textbf{4.54} & 0.83 & 0.80 & 0.78 & 0.77 & 0.77 & 0.76 & 0.76 \\
\hline
\end{tabular}
\caption{Effect of the number of MTP steps during training with \hsm on
\lightning{} and \super3dot5{}, evaluated at $T=1$ with probabilistic drafting
and draft length seven. Pre-boosting denotes the \mtp module before boosting.}
\label{tab:mtp_recurrence_depth_t1}
\end{table}

\newpage
\FloatBarrier

\subsection{Training Cost of Additional MTP Steps in MTP Boosting}
\label{app:mtp-training-cost}

We measure training iteration time for \super{} with a frozen
backbone and two, three, five, or seven recurrent \mtp steps. All runs use
shared \mtp parameters, uniform layer sampling for \hsm, and KL
distillation. Each run uses GB300 GPUs, BF16 precision, a global batch size of
64, a microbatch size of one, and a sequence length of 8k. Backbone hidden states are recomputed
on each iteration rather than cached.
The frozen-backbone configurations use the same allocation and parallelism
settings. Each configuration runs for 100 iterations. We exclude the first ten
iterations and compute the mean iteration time over the remaining 90.
Table~\ref{tab:mtp-training-cost} summarizes the iteration-time increase relative to a
two MTP step configuration.

\begin{table}[!htb]
\centering
  \setlength{\tabcolsep}{3pt}
\small
\begin{tabular}{@{}lcr@{}}
\toprule
Backbone & \mtp steps & Iteration-time increase \\
\midrule
\multirow{4}{*}{Frozen} & 2 & 0.0\% \\
 & 3 & 7.8\% \\
 & 5 & 22.6\% \\
 & 7 & 45.2\% \\
\bottomrule
\end{tabular}
\caption{Relative increase in mean training iteration time on \super{},
measured against a two MTP step setting.}
\label{tab:mtp-training-cost}
\end{table}

\subsection{KL Loss Temperature}
\label{app:kl-loss-temperature}

We isolate the effect of KL loss temperature on \super{} using
two \mtp training steps, uniform layer sampling for \hsm, and a
frozen backbone. All three runs use the same on-policy V2 data generated
at $\tau=1$ and are trained for 4k iterations with sequence length
8k, global batch size 64, and learning rate $10^{-5}$. We vary only the loss temperature
$T_{\mathrm{KL}}\in\{1,2,3\}$, applied to both target and draft logits,
with the KL objective multiplied by $T_{\mathrm{KL}}^2$ following \cite{hinton2015distilling}.

\begin{figure}[!htb]
\centering
\begin{minipage}[t]{0.55\linewidth}
  \vspace{0pt}
  \begin{minipage}[c][4.6cm][c]{\linewidth}
    \centering
    \footnotesize
    \setlength{\tabcolsep}{1.6pt}
    \renewcommand{\arraystretch}{1.15}
\begin{tabular}{@{}c*{8}{c}@{}}
\toprule
$T_{\mathrm{KL}}$ & AL & AR@1 & AR@2 & AR@3 & AR@4 & AR@5 & AR@6 & AR@7 \\
\midrule
\multicolumn{9}{c}{Evaluation $T=0$, greedy drafting} \\
\midrule
1 & 3.69 & \textbf{0.83} & \textbf{0.78} & 0.72 & 0.64 & 0.54 & 0.47 & 0.39 \\
2 & \textbf{3.78} & \textbf{0.83} & \textbf{0.78} & \textbf{0.73} & 0.65 & 0.58 & 0.51 & 0.47 \\
3 & 3.73 & 0.82 & 0.77 & 0.72 & \textbf{0.66} & \textbf{0.60} & \textbf{0.56} & \textbf{0.52} \\
\midrule
\multicolumn{9}{c}{Evaluation $T=1$, probabilistic drafting} \\
\midrule
1 & 3.36 & 0.80 & 0.75 & 0.66 & 0.56 & 0.45 & 0.37 & 0.30 \\
2 & \textbf{3.64} & \textbf{0.82} & \textbf{0.77} & \textbf{0.71} & 0.63 & 0.55 & 0.48 & 0.42 \\
3 & 3.62 & 0.81 & 0.76 & 0.70 & \textbf{0.64} & \textbf{0.58} & \textbf{0.53} & \textbf{0.49} \\
\bottomrule
\end{tabular}
  \end{minipage}
  \captionof{table}{KL loss-temperature ablation on \super{} after 4k
  iterations with two shared \mtp steps and \hsm. Bold marks the highest
  value for each metric within each evaluation temperature.}
  \label{tab:super-kl-temperature}
\end{minipage}\hfill
\begin{minipage}[t]{0.43\linewidth}
  \vspace{0pt}
  \begin{minipage}[c][4.6cm][c]{\linewidth}
    \centering
    \includegraphics[width=\linewidth]{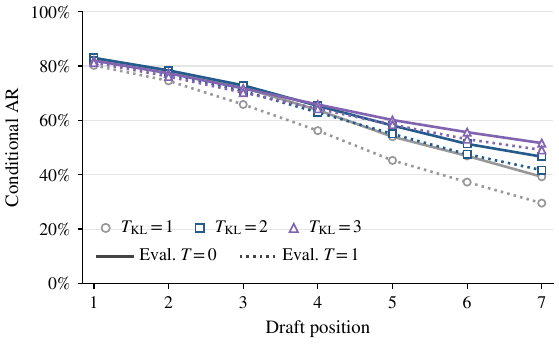}
  \end{minipage}
  \caption{Acceptance rates for Table~\ref{tab:super-kl-temperature}.
  Colors denote loss temperature $T_{\mathrm{KL}}$; solid and dotted lines
  denote greedy and probabilistic ($T=1$) evaluation, respectively.}
  \label{fig:super-kl-temperature}
\end{minipage}
\end{figure}

Table~\ref{tab:super-kl-temperature} reports AL and acceptance rates on SPEED-Bench qualitative split. All six
evaluations use vLLM, draft length seven, and the same serving
configuration. Evaluation temperature $T=0$ uses greedy drafting;
$T=1$ uses probabilistic drafting. Evaluation temperature is distinct
from both $T_{\mathrm{KL}}$ and the rollout generation temperature $\tau$.

Increasing $T_{\mathrm{KL}}$ from 1 to 2 raises AL from 3.69 to 3.78 at
$T=0$ and from 3.36 to 3.64 at $T=1$. Increasing it further to 3 improves
acceptance rates at positions four through seven under both
evaluation temperatures, but reduces acceptance rates at the first three
positions (Figure~\ref{fig:super-kl-temperature}). Consequently,
$T_{\mathrm{KL}}=2$ achieves the highest overall AL among the tested
settings. 

\FloatBarrier

\par

\FloatBarrier

\newpage
\subsection{EAGLE-3 Acceptance Rates at Temperature One}
\label{app:eagle-ar-t1}

\begin{figure}[!htb]
  \centering
  \includegraphics[width=\linewidth]{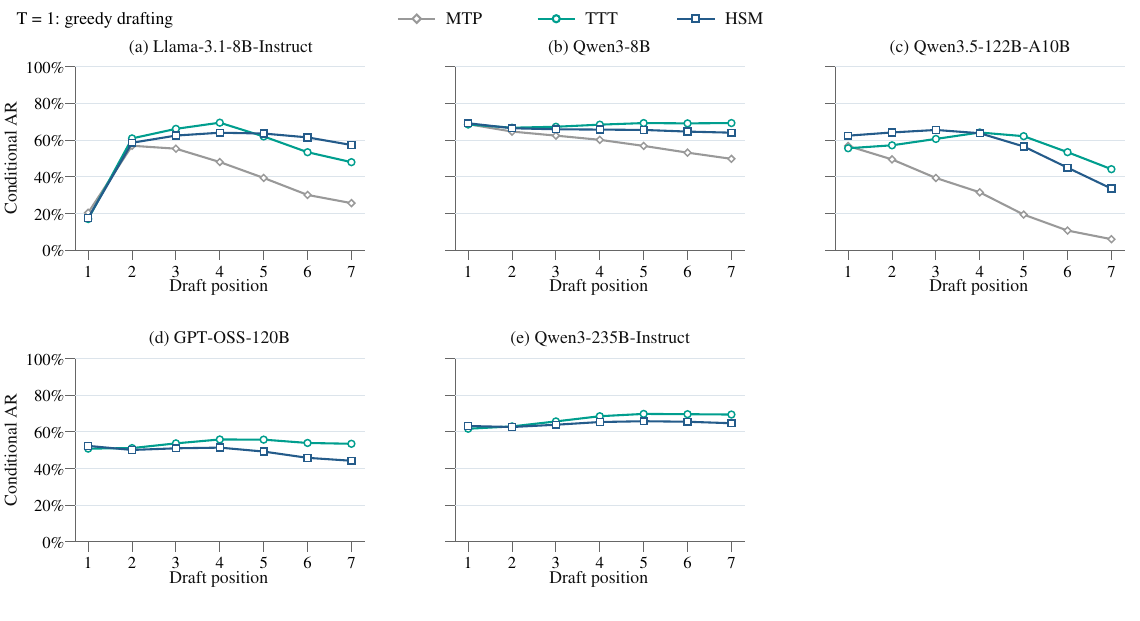}
  \caption{Acceptance rates at $T=1$ with greedy drafting on the SPEED-Bench
  qualitative split, using draft length seven. }
  \label{fig:eagle-ar-t1-greedy}
\end{figure}

\begin{figure}[!htb]
  \centering
  \includegraphics[width=\linewidth]{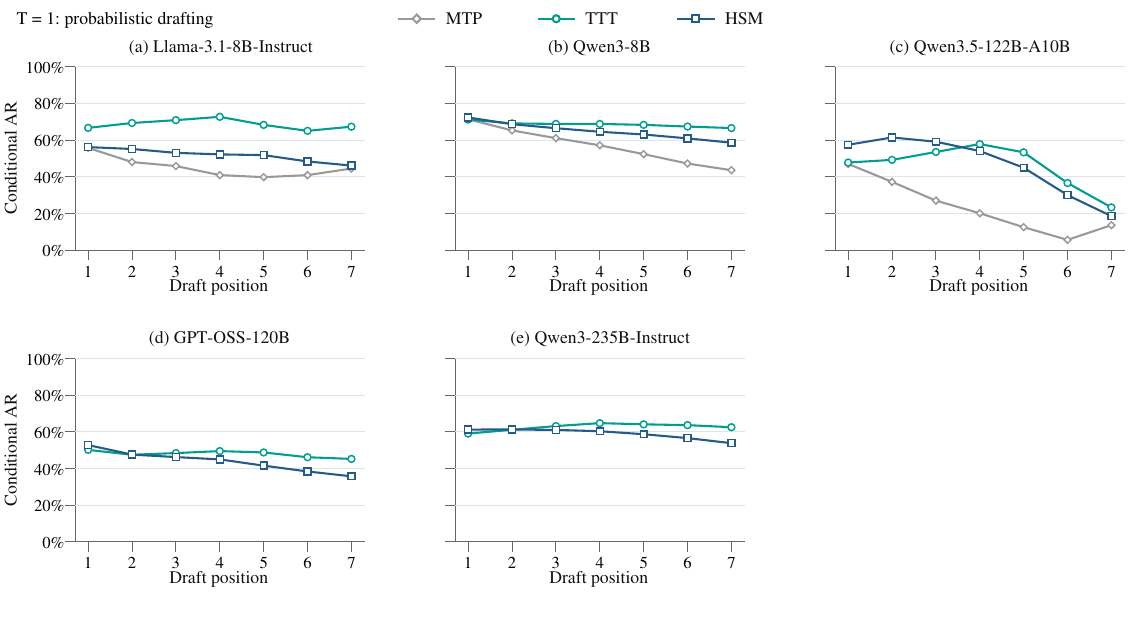}
  \caption{Acceptance rates at $T=1$ with probabilistic drafting on the
  SPEED-Bench qualitative split, using draft length seven. }
  \label{fig:eagle-ar-t1-probabilistic}
\end{figure}

Figures~\ref{fig:eagle-ar-t1-greedy} and~\ref{fig:eagle-ar-t1-probabilistic}
report acceptance rates at $T=1$ with greedy and probabilistic drafting,
respectively. Both use the five target models and selected checkpoints
in Table~\ref{tab:eagle_vs_hsm_speed}, evaluated at draft length seven.

\FloatBarrier

\end{document}